\documentclass{article}

 \usepackage[preprint]{neurips_2026}

\usepackage[utf8]{inputenc} 
\usepackage[T1]{fontenc}    
\usepackage{url}            
\usepackage{booktabs}       
\usepackage{amsfonts}       
\usepackage{nicefrac}       
\usepackage{microtype}      

\usepackage{orcidlink}

\usepackage{bbding}
\usepackage{multirow}      
\usepackage{xcolor}        
\usepackage{caption}       
\usepackage{subcaption}    
\usepackage{inconsolata}
\usepackage{colortbl}
\usepackage{makecell}
\usepackage{tabulary}
\usepackage{fontawesome5}
\usepackage{pifont}
\usepackage{tikz}
\usepackage{amsfonts}
\usepackage{bm}
\usepackage{algorithm}
\usepackage{algorithmicx}
\usepackage{algpseudocode}
\usepackage{wrapfig}
\usepackage{adjustbox}
\usepackage{booktabs} 
\usepackage{graphicx}

\definecolor{citecolor}{HTML}{0071bc} 
\hypersetup{
    colorlinks=true,
    linkcolor=red,
    citecolor=citecolor,
    urlcolor=citecolor,
    pdfborder={0 0 0},
    breaklinks=true
}

\usepackage{marvosym}
\newcommand{\blfootnote}[1]{%
  \begingroup
  \renewcommand\thefootnote{}\footnote{#1}%
  \addtocounter{footnote}{-1}%
  \endgroup
}

\newcommand{\conf}[1]{{\!\tiny\texttt{(#1)}}}

\usepackage[accsupp]{axessibility}  
\title{V-CAST: Video Curvature-Aware Spatio-Temporal Pruning for Efficient Video Large Language Models}

\author{
Xinying Lin$^{1,2}$ \quad
Xuyang Liu$^{3\dagger}$ \quad
Yiyu Wang$^{4}$ \quad
Teng Ma$^{1}$ \quad
Wenqi Ren$^{1,2}$$^{\text{\Envelope}}$
\\
\\
$^1$Shenzhen Campus of Sun Yat-sen University \quad
$^2$Shenzhen Loop Area Institute \\
$^3$Sichuan University \quad
$^4$EPIC Lab, Shanghai Jiao Tong University \\
\vspace{-6pt}\\
\textbf{Homepage:} \url{https://xinyouu.github.io/V-CAST/}
}

\makeatletter
\let\@oldmaketitle\@maketitle
\renewcommand{\@maketitle}{\@oldmaketitle
\centering
    \centering
    \vspace{-4mm}
    \includegraphics[width=\textwidth]{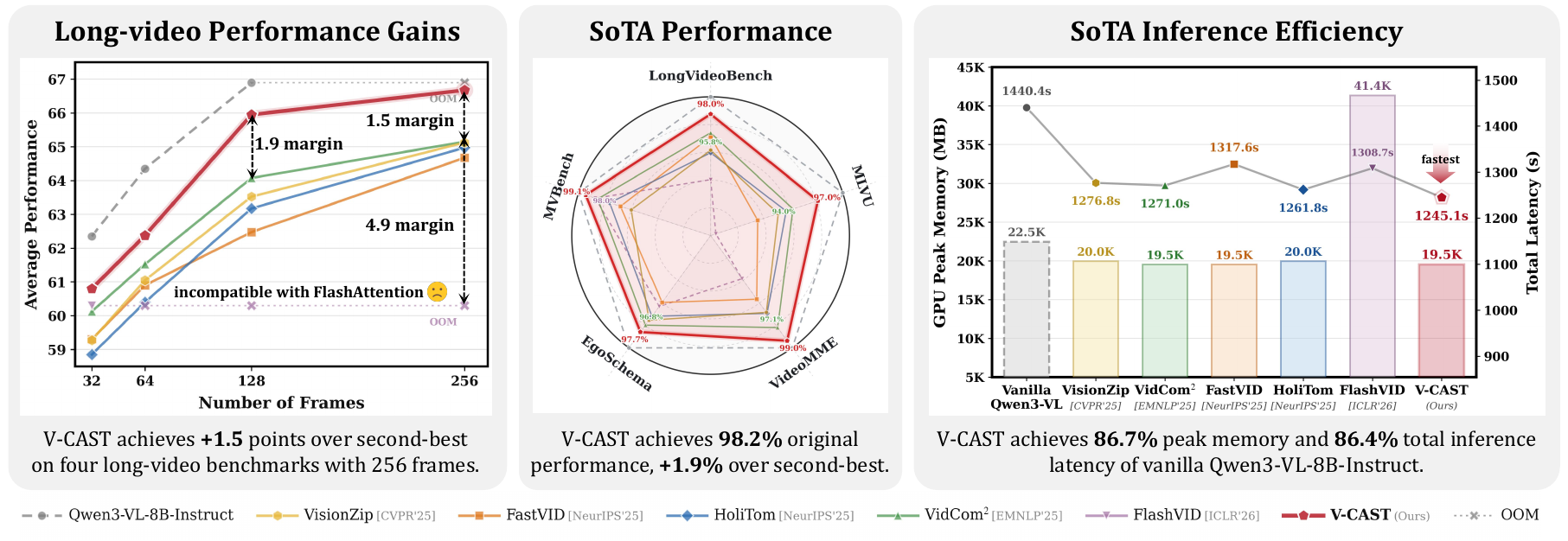}
    \vspace{-6mm}
    \captionof{figure}{\textbf{Highlight of V-CAST.} Compared to existing token compression methods~\cite{yang2025visionzip,shen2025fastvid,shao2025holitom,liu2025vidcom2,fan2026flashvid}, V-CAST yields \textbf{+1.5} long-video gains, preserves \textbf{98.2\%} original performance with a \textbf{+1.9\%} margin, and reduces peak memory and latency to \textbf{86.7\%} and \textbf{86.4\%} of Qwen3-VL-8B-Instruct~\cite{Qwen3-VL}.}
    \vspace{-3mm}
\label{fig:teaser}
}

\begin{document}
\maketitle

\blfootnote{$\dagger$ Project leader: \texttt{seanleo666@gmail.com}. $^{\text{\Envelope}}$ Corresponding author: \texttt{renwq3@mail.sysu.edu.cn}.}

\begin{abstract}
\vspace{-0.2cm}
Video large language models (VideoLLMs) show strong capability in video understanding, yet long-context inference is still dominated by massive redundant visual tokens in the prefill stage. 
We revisit token compression for VideoLLMs under a tight budget and identify a key bottleneck, namely insufficient \textit{spatio-temporal information coverage}. Existing methods often introduce discontinuous coverage through coarse per-frame allocation or scene segmentation, and token merging can further misalign spatio-temporal coordinates under MRoPE-style discrete $(t,h,w)$ bindings. To address these issues, we propose V-CAST (\textbf{V}ideo \textbf{C}urvature-\textbf{A}ware \textbf{S}patio-\textbf{T}emporal Pruning), a training-free, plug-and-play pruning policy for long-context video inference. V-CAST casts token compression as a trajectory approximation problem and introduces a curvature-guided temporal allocation module that routes per-frame token budgets to semantic turns and event boundaries. It further adopts a dual-anchor spatial selection mechanism that preserves high-entropy visual evidence without attention intervention, while keeping retained tokens at their original coordinates to maintain positional alignment. Extensive experiments across multiple VideoLLMs of different architectures and scales demonstrate that V-CAST achieves \textbf{98.6\%} of the original performance, outperforms the second-best method by \textbf{+1.1\%} on average, and reduces peak memory and total latency to \textbf{86.7\%} and \textbf{86.4\%} of vanilla Qwen3-VL-8B-Instruct. 
\end{abstract}    
\section{Introduction}
\label{sec:intro}

Video large language models (VideoLLMs)~\cite{wang2025internvideo2.5,chen2024longvila,zhang2025videollama,yang2025longvt,yang2025timeexpert} extend large vision-language models (LVLMs) to large-scale video data~\cite{li2024llava-ov,zhang2024llava-video} and have achieved strong performance on diverse video understanding tasks~\cite{fu2024videomme,wen2025Alpha-Service,wang2025stc,tao2026lvomnibench}.
However, videos introduce highly dense tokens from consecutive frames, so long-context prefill becomes the dominant compute and memory cost in attention-based inference~\cite{vaswani2017attention}. 
For example, advanced VideoLLM Qwen3-VL~\cite{Qwen3-VL} supports up to a 256K context window and performs well on long-video understanding~\cite{wang2025lvbench,wu2024longvideobench,zhou2024mlvu}, but scaling video inputs to such long contexts becomes prohibitively expensive, which limits practical deployment~\cite{liu2025shifting,shao2025survey}.

To alleviate this efficiency bottleneck, token compression reduces visual tokens by pruning redundant content~\cite{han2026ficoco,huang2024prunevid,chen2025v2drop,tao2024dycoke,liu2026globalcom2,liu2025mixkv,yuan2025unicomp,ji2025specvlm,ding2026omnisift,tao2025omnizip,kong2026parallelvlm,xing2024pdrop}. 
Recent works further tailor compression to videos by introducing frame-wise control, including \textit{scene-based segmentation} pipelines~\cite{shen2025fastvid,shao2025holitom} and \textit{frame-wise token budgeting} strategies~\cite{liu2025video,ma2025mmg-vid}.
Despite these advances, under a fixed and tight budget it remains difficult to preserve \textbf{spatio-temporal information coverage}, as reflected by the budgeting profiles in Figure~\ref{fig:budget_analysis}. We find that this limitation mainly stems from \textbf{\textit{two key factors}}:

\begin{figure*}[!t]
    \centering
 \includegraphics[width=\linewidth]{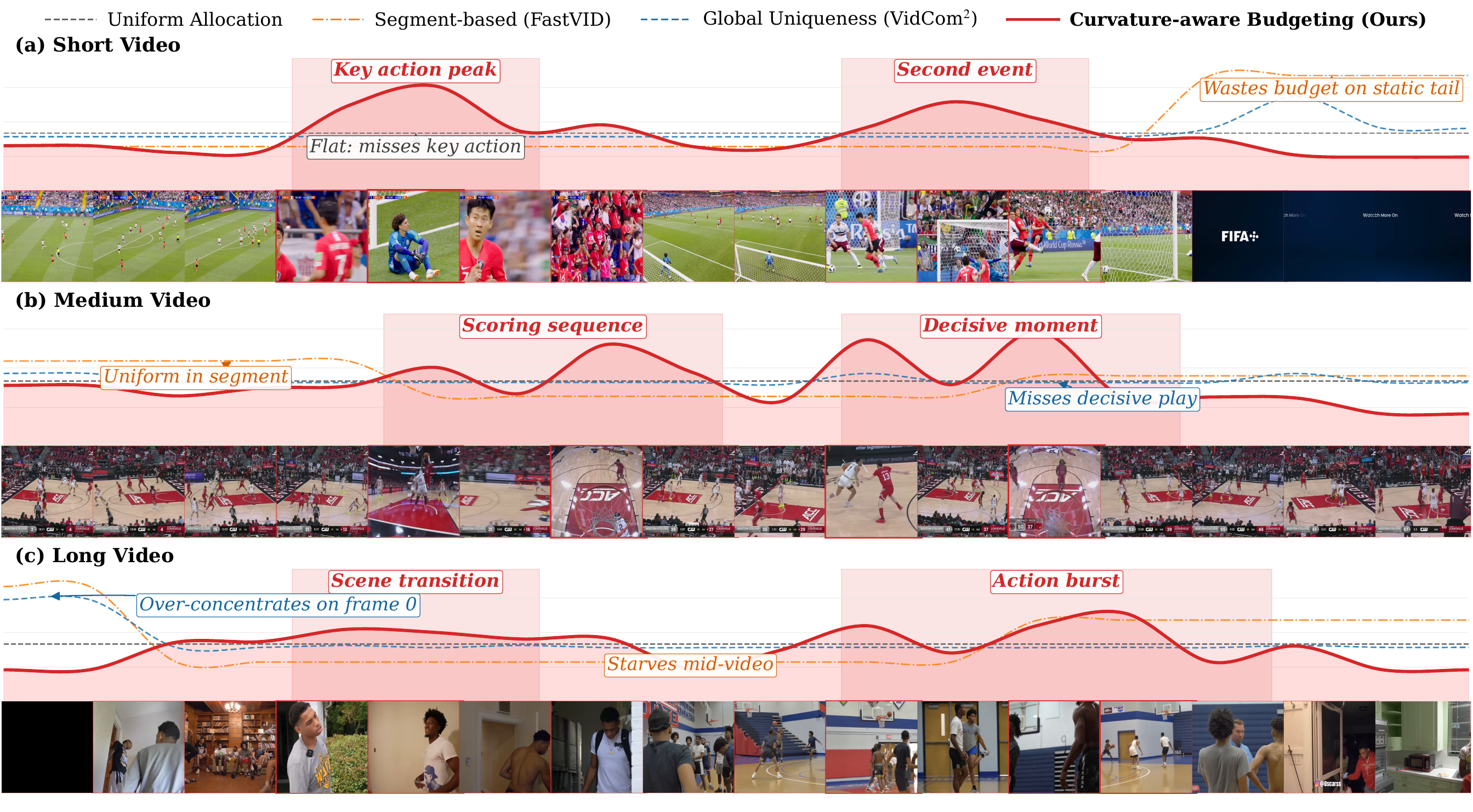}
    \vspace{-6mm} 
    \caption{\textbf{Spatio-temporal budgeting profiles.} We visualize per-frame token budgets (higher means more retained tokens) on representative short, medium, and long videos. \textcolor[rgb]{0.5, 0.5, 0.5}{Uniform Allocation}~\cite{yang2025visionzip} stays nearly flat and misses brief evidence peaks; \textcolor[rgb]{0.8, 0.4, 0.1}{segment-based} pinelines~\cite{shen2025fastvid,shao2025holitom} introduce boundary-sensitive budget jumps and waste tokens within redundant windows; \textcolor[rgb]{0.2, 0.4, 0.7}{global-uniqueness} budgeting~\cite{liu2025vidcom2} can over-concentrate on a few globally distinctive frames and under-cover transitional segments. \textcolor[rgb]{0.780,0.180,0.220}{\textbf{Curvature-aware}} budgeting allocates more tokens to rapid semantic changes and event boundaries, improving spatiotemporal information coverage under tight budgets.}
    \vspace{-4mm}
    \label{fig:budget_analysis}
\end{figure*}

\noindent \textbf{(I) Discontinuous Spatio-temporal Information:} In human video perception, attention is distributed unevenly over time, yet understanding relies on continuous context because decisive moments are supported by nearby transitions. Uniform compression~\cite{yang2025visionzip} keeps an identical budget for every frame; the flat profile in Figure~\ref{fig:budget_analysis} shows that it cannot upweight brief evidence peaks, so critical moments are easily diluted under tight budgets. Segment-based pipelines (\textit{e.g.}, FastVID~\cite{shen2025fastvid} and HoliTom~\cite{shao2025holitom}) partition videos and compress tokens within each segment, inducing boundary-sensitive budget jumps and near-uniform allocation inside segments, weakening continuity across transitions. Global-uniqueness budgeting such as VidCom$^2$~\cite{liu2025vidcom2} allocates budgets by global frame distinctiveness, which can over-concentrate on a few globally salient but not locally decisive frames, leaving transitional segments under-covered and causing discontinuous coverage under tight budgets.

\begin{wrapfigure}{r}{0.52\textwidth}
    \centering
    \vspace{-1.2em}
    \captionof{table}{Comparisons of different spatial selection strategies at $R=25\%$.}
    \vspace{-0.3em}
    \label{tab:rope}
    \resizebox{\linewidth}{!}{
        \begin{tabular}{lcccccc}
        \toprule
        \footnotesize
        \multirow{2}{*}{\textbf{Spatial Selection}} &
        \multirow{2}{*}{\textbf{MVBench}} &
        \multirow{2}{*}{\makecell{\textbf{LongVideo}\\\textbf{Bench}}} &
        \multicolumn{4}{c}{\textbf{VideoMME}} \\
        \cmidrule(lr){4-7}
            &   &  & \textbf{All} & \textbf{S} & \textbf{M} & \textbf{L} \\
        \midrule
        \textcolor[rgb]{ .502, .502, .502}{Qwen3-VL-8B-Instruct} &
        \textcolor[rgb]{ .502, .502, .502}{68.6} &
        \textcolor[rgb]{ .502, .502, .502}{60.3} &
        \textcolor[rgb]{ .502, .502, .502}{64.5} &
        \textcolor[rgb]{ .502, .502, .502}{76.0} &
        \textcolor[rgb]{ .502, .502, .502}{60.4} &
        \textcolor[rgb]{ .502, .502, .502}{57.0} \\
        \midrule
        \multicolumn{7}{l}{\textbf{(a) Token Merging}} \\
        HoliTom \conf{NeurIPS'25} &63.0 &56.8 &59.7 &71.4 &54.6 &53.1 \\
        AvgPool ($2\times2$) & 65.4 & 58.6 & 60.5 & 72.3 & 56.7 & 52.4 \\
        \midrule
        \multicolumn{7}{l}{\textbf{(b) Token Pruning}} \\
        Random ($2\times2$) & 66.9 & 57.8 & 61.9 & 73.6 & 58.7 & 53.4 \\
        First ($2\times2$) & 67.7 & 57.7  & 61.1 & 72.2 & 56.2 & 55.0 \\
        
        \bottomrule
        \end{tabular}
    }
    \vspace{0.6em}
    \includegraphics[width=\linewidth]{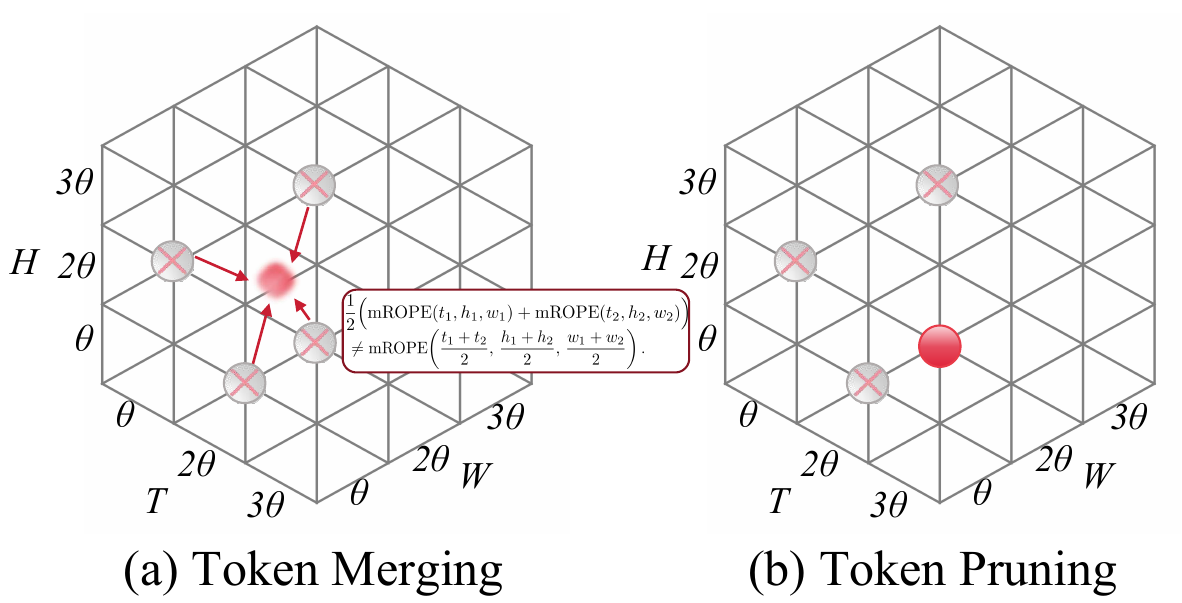}
    \vspace{-7.2mm}
    \captionof{figure}{Token merging drifts off the discrete $(t,h,w)$ grid and and weakens MRoPE~\cite{Qwen3-VL} bindings, while token pruning preserves on-grid coordinates.}
    \label{fig:rope}
    \vspace{-2.2em}
\end{wrapfigure}

\noindent \textbf{(II) Misaligned Spatio-temporal Information:} Advanced VideoLLMs (\textit{e.g.}, Qwen3-VL~\cite{Qwen3-VL}) adopt MRoPE to tie each visual token to discrete $(t,h,w)$ coordinates for spatio-temporal reasoning.
Token merging (\textit{i.e.}, Merging) aggregates multiple tokens into one (\textit{e.g.}, average pooling), which shifts the merged token off the original grid and breaks the one-to-one correspondence with MRoPE in this setting, as illustrated in Figure~\ref{fig:rope}. 
This effect is reflected in Table~\ref{tab:rope}: the carefully designed HoliTom~\cite{shao2025holitom} even underperforms the naive AvgPool baseline by \textbf{2.4} points on MVBench~\cite{li2024mvbench}, \textbf{1.8} on LongVideoBench~\cite{wu2024longvideobench}, and \textbf{0.8} on VideoMME~\cite{fu2024videomme}. 
Notably, a simple pruning baseline that keeps one token per $2\times2$ cell already surpasses merging by a large margin, improving over HoliTom by \textbf{+2.2} on VideoMME.
In contrast, token pruning selects existing tokens and therefore preserves on-grid coordinates, maintaining positional alignment under MRoPE.

Based on the above analysis, token compression for VideoLLMs can be formulated as the following question: \textit{how can we select a compact set of visual tokens under a fixed and tight budget while preserving continuous and well-aligned spatio-temporal information coverage?}.

To answer this question, we propose \textbf{V}ideo \textbf{C}urvature-\textbf{A}ware \textbf{S}patio-\textbf{T}emporal Pruning (\textit{i.e.}, \textbf{V-CAST}), a plug-and-play token pruning approach from the perspective of \textit{video spatio-temporal curvature}. V-CAST couples \textit{per-frame token budgeting} to allocate capacity to informative temporal transitions with \textit{content-driven token selection} to further preserve diverse spatial evidence, while naturally keeping retained tokens at their original coordinates to maintain positional alignment.

As highlighted in Figure~\ref{fig:teaser}, V-CAST achieves a strong accuracy--efficiency trade-off for long-context video inference, improving long-video understanding while reducing memory and latency. Specifically, it yields \textbf{+1.5} points over the second-best method on four long-video benchmarks with 256 frames, achieves \textbf{98.2\%} original performance with a \textbf{+1.9\%} margin, and improves efficiency to \textbf{86.7\%} peak memory and \textbf{86.4\%} total inference latency of vanilla Qwen3-VL-8B-Instruct.

In summary, our contributions are threefold:

\begin{enumerate}
    \item \textbf{Spatio-temporal Coverage Rethinking.} We revisit visual token compression for VideoLLMs under tight token budgets and identify \textit{spatio-temporal information coverage} as the core bottleneck in video understanding. We further reveal two failure modes in prior pipelines: \textit{discontinuous} coverage from segmentation or coarse budgeting, and \textit{misaligned} spatio-temporal information from token merging under MRoPE.

    \item \textbf{Video Curvature-Aware Spatio-Temporal Pruning.} Based on these findings, we propose \textbf{V-CAST}, a plug-and-play token pruning approach that couples \textit{per-frame token budgeting} with \textit{content-driven token selection} from the perspective of video spatio-temporal curvature, while keeping retained tokens at their original coordinates to preserve alignment.

    \item \textbf{State-of-the-Art Performance and Efficiency.} Experiments on long-video benchmarks show that V-CAST achieves the best accuracy--efficiency trade-off on Qwen3-VL-8B/32B-Instruct and LLaVA-OV/Video-7B. Under 128-frame inference, V-CAST exceeds the second-best method by \textbf{+1.9} points on average, while reducing peak memory and end-to-end latency.
\end{enumerate}

\section{Related Work}

\noindent \textbf{Video Large Language Models.} Large vision-language models (LVLMs) commonly follow the ``ViT--Projector--LLM'' paradigm, where visual tokens are projected into the LLM embedding space for effective autoregressive multimodal reasoning~\cite{Liu:LLaVA-1.5,Bai:Qwen-VL,Chen:InternVL-v1.5}. 
To support video understanding, recent video large language models (VideoLLMs) extend this framework with temporal modeling and training strategies~\cite{zhang2024llava-video,shu2024video-xl}. 
LLaVA-OneVision~\cite{li2024llava-ov} unifies image and video inputs in a single model, while LLaVA-Video~\cite{zhang2024llava-video} further scales video instruction data to strengthen temporal understanding.
To improve spatio-temporal reasoning, Qwen2.5-VL~\cite{bai2025qwen2} and Keye-VL~\cite{team2025kwai} adopt Multimodal Rotary Positional Embedding (M-RoPE), which injects temporal and spatial coordinates into rotary embeddings so tokens remain aware of their $(t,h,w)$ positions.
Qwen3-VL~\cite{Qwen3-VL} scales this capability with longer context and high-quality improved data, while InternVL3.5~\cite{wang2025internvl3} leverages video-centric training to model complex dynamics. 
Compared with images, video understanding processes substantially more visual tokens and typically exhibits higher spatio-temporal redundancy, which makes real-time video understanding difficult to achieve efficiently in practice.

\noindent \textbf{Token Compression for VideoLLMs.}
Token compression reduces computation by pruning redundant visual tokens and is often training-free and plug-and-play at the inference stage~\cite{Chen:FastV,yang2025visionzip}. For VideoLLMs, video-specific compression methods mainly control tokens along the temporal axis in two representative paradigms. \textbf{(i)} \textit{Scene-based compression} first partitions a video into temporal windows and then compresses tokens within each window. FastVID~\cite{shen2025fastvid} prunes tokens based on intra-window redundancy, while HoliTom~\cite{shao2025holitom} combines segmentation with spatio-temporal token merging. These pipelines introduce an explicit segmentation step and can be sensitive to boundary quality, especially under gradual transitions or subtle but task-critical changes. \textbf{(ii)} \textit{Budget-based compression} allocates per-frame token budgets before selecting tokens. VidCom$^2$~\cite{liu2025vidcom2} adjusts compression intensity based on frame uniqueness, but a global distinctiveness signal can over-concentrate budgets on a few frames and under-cover transitional segments, which hurts temporal continuity under aggressive budgets. MMG-Vid~\cite{ma2025mmg-vid} combines segmentation and budgeting, but it still retains an explicit segmentation stage and additional components. As a result, existing methods can produce discontinuous or misaligned spatio-temporal coverage under tight budgets, leading to suboptimal accuracy in practical scenarios. Motivated by these limitations, V-CAST coordinates temporal budgeting with spatial token selection to improve spatio-temporal coverage under aggressive budgets.

\section{Methodology}

\subsection{Preliminary}
\label{sec:preliminary}

\noindent \textbf{VideoLLM Pipeline.}
Standard VideoLLMs follow the ``ViT--Projector--LLM'' paradigm with three stages. 
\textbf{(i)} \textit{Vision encoding.} Given a video $\mathcal{V}\in\mathbb{R}^{T\times H\times W\times 3}$ with $T$ frames, height $H$, and width $W$, a vision encoder $\phi(\cdot)$ produces patch features $\mathbf{Z}=\phi(\mathcal{V})\in\mathbb{R}^{T\times P\times D}$, where $P$ is the number of patch tokens per frame and $D$ is the feature dimension. A projector $g(\cdot)$ maps them into the LLM feature space, yielding visual tokens $\mathbf{X}=g(\mathbf{Z})\in\mathbb{R}^{T\times P\times D'}$ with token dimension $D'$. Flattening all frames gives $\mathbf{X}^{\flat}\in\mathbb{R}^{N\times D'}$ with $N=T\cdot P$ visual tokens. 
\textbf{(ii)} \textit{LLM pre-filling.} Let $\mathbf{Y}\in\mathbb{R}^{M\times D'}$ denote $M$ text tokens. The LLM processes the concatenated sequence $\mathbf{S}_0=[\mathbf{X}^{\flat};\mathbf{Y}]\in\mathbb{R}^{(N+M)\times D'}$ and caches layer-wise key-value (KV) pairs $\{(\mathbf{K}^{\ell},\mathbf{V}^{\ell})\}_{\ell=1}^{L}$ across $L$ layers. 
\textbf{(iii)} \textit{LLM decoding.} At step $s$, the model attends to cached KV pairs and generates $p(y_s\mid \mathcal{V},y_{<s})=\mathrm{Softmax}(\mathbf{W}\mathbf{h}_s)$, where $\mathbf{h}_s=\mathrm{Attn}(\mathbf{q}_s;\mathbf{K}_{\mathrm{cache}},\mathbf{V}_{\mathrm{cache}})$. Repeating this auto-regressively yields the final response.

\noindent \textbf{Efficiency Bottleneck.}
Compared with images, videos contain a substantially larger number of visual tokens since $N=T\cdot P$ grows with $T$. In pre-filling, self-attention over $\mathbf{S}_0$ incurs quadratic complexity $\mathcal{O}((N+M)^2)$, resulting in longer latency and higher peak GPU memory. Meanwhile, the KV cache at layer $\ell$ stores $\mathbf{K}^{\ell},\mathbf{V}^{\ell}\in\mathbb{R}^{(N+M)\times d}$ (with head dimension $d$), so the total cache memory scales as $\mathcal{O}(L(N+M)d)$. This also increases decoding latency and GPU memory footprint, making VideoLLM inference significantly more expensive than image understanding.

\subsection{V-CAST: Curvature-Aware Spatio-Temporal Pruning}

\noindent \textbf{Problem Formulation.}
In VideoLLMs, the visual sequence length scales as $N=T\cdot P$, which makes the pre-filling attention over $\mathbf{S}_0=[\mathbf{X}^{\flat};\mathbf{Y}]$ expensive.
Token compression therefore aims to construct a shorter visual sequence $\tilde{\mathbf{X}}^{\flat}$ under a strict budget $B=\lfloor rN \rfloor$ ($r\in(0,1)$), seeking to retain downstream video understanding performance.
Concretely, we select an index set $\mathcal{I}\subseteq\{1,\ldots,N\}$ with $|\mathcal{I}|=B$ and keep the original order, yielding $\tilde{\mathbf{X}}^{\flat}=\mathbf{X}^{\flat}[\mathcal{I}]$.
Under tight budgets, the key challenge is to preserve continuous and well-aligned spatio-temporal information coverage rather than allocating tokens uniformly in time or pooling them statically in space.

\begin{figure*}[!t]
    \centering
 \includegraphics[width=\linewidth]{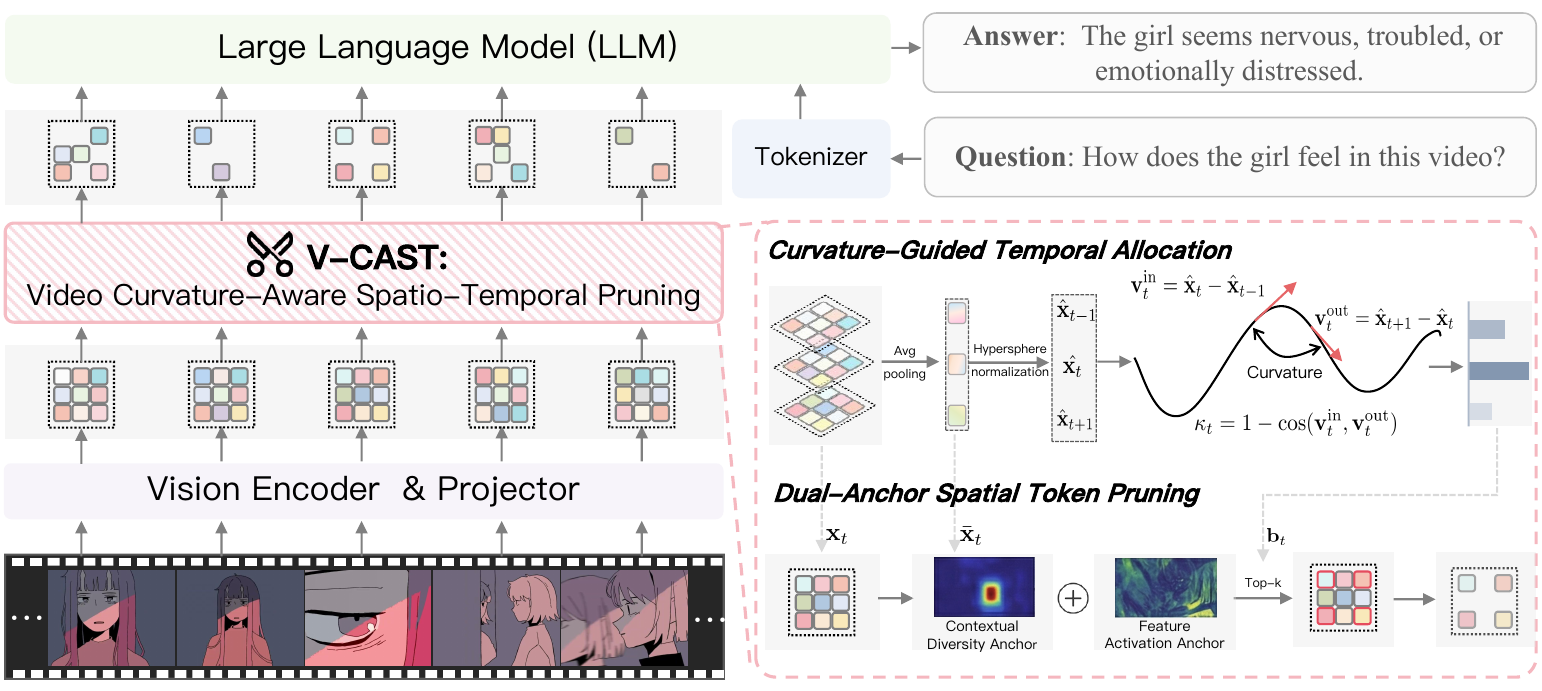}
    \vspace{-4mm} 
    \caption{\textbf{Overview of V-CAST.} V-CAST formulates token pruning for VideoLLMs as an optimal semantic-trajectory approximation problem under a fixed budget. It applies \textbf{Curvature-Guided Temporal Allocation} to assign per-frame budgets by tracking semantic transitions, and then performs \textbf{Dual-Anchor Spatial Token Selection} to retain diverse and salient tokens within each frame while preserving their original on-grid coordinates.}
    \label{fig:overview}
    \vspace{-2mm}
\end{figure*}

\noindent \textbf{Overview.}
Our design is guided by two questions:
\textbf{(Q1)} \textit{When should we allocate tokens across frames under a fixed budget?}.
\textbf{(Q2)} \textit{Where should we keep tokens within each frame to preserve diverse visual evidence?}.
To answer them, we propose \textbf{V-CAST} (Figure~\ref{fig:overview}), a training-free and plug-and-play pruning framework with two components.
\textbf{(i) Curvature-Guided Temporal Allocation} (Section~\ref{subsec:temporal_budget}) rationally distributes budgets over time based on semantic transitions  across frames, and \textbf{(ii) Dual-Anchor Spatial Token Selection} (Section~\ref{subsec:spatial_sampling}) retains a compact yet informative set of tokens while preserving the on-grid coordinates.

\subsection{Curvature-Guided Temporal Allocation}
\label{subsec:temporal_budget}

We allocate the token budget across frames by measuring how the video semantics evolves over time.
Let $\mathbf{x}_{t,i}\in\mathbb{R}^{D'}$ denote the projected visual token at frame $t$ and spatial index $i$.
We first summarize each frame into a mean semantic vector
\begin{equation}
\bar{\mathbf{x}}_t=\frac{1}{P}\sum_{i=1}^{P}\mathbf{x}_{t,i}\in\mathbb{R}^{D'},
\end{equation}
which preserves both direction and magnitude of the frame representation.
We then normalize it to a unit direction
\begin{equation}
\hat{\mathbf{x}}_t=\frac{\bar{\mathbf{x}}_t}{\|\bar{\mathbf{x}}_t\|_2}\in\mathbb{S}^{D'-1},
\end{equation}
so each frame is represented as a point on the hypersphere, capturing its semantic direction.
Connecting $\{\hat{\mathbf{x}}_t\}_{t=1}^{T}$ in temporal order forms a semantic trajectory on $\mathbb{S}^{D'-1}$, where smooth segments indicate redundant content and sharp turns often correspond to salient transitions.

For boundary frames, we use one-sided differences by clamping indices, \textit{i.e.}, $\hat{\mathbf{x}}_{0}=\hat{\mathbf{x}}_{1}$ and $\hat{\mathbf{x}}_{T+1}=\hat{\mathbf{x}}_{T}$.
We define the incoming and outgoing directions at frame $t$ as
\begin{equation}
\mathbf{v}_{t}^{\text{in}}=\hat{\mathbf{x}}_t-\hat{\mathbf{x}}_{t-1}, \quad
\mathbf{v}_{t}^{\text{out}}=\hat{\mathbf{x}}_{t+1}-\hat{\mathbf{x}}_t,
\end{equation}
and measure the semantic curvature by the change of directions:
\begin{equation}
\kappa_t = 1 - \cos(\mathbf{v}_{t}^{\text{in}}, \mathbf{v}_{t}^{\text{out}}),
\end{equation}
where $\cos(\mathbf{a},\mathbf{b})=\frac{\mathbf{a}^{\top}\mathbf{b}}{\|\mathbf{a}\|_2\|\mathbf{b}\|_2+\epsilon}$ and $\epsilon$ is a small constant.

Given the global budget $B$, we allocate a budget density $p_t$ over frames by solving an entropy-regularized allocation problem:
\begin{equation}
\max_{\mathbf{p}\in\Delta^{T-1}} \sum_{t=1}^{T} p_t \kappa_t + \tau \mathcal{H}(\mathbf{p}), \quad
\mathcal{H}(\mathbf{p})=-\sum_{t=1}^{T} p_t\log p_t,
\end{equation}
where $\Delta^{T-1}$ is the probability simplex and $\tau$ is a temperature hyperparameter controlling allocation smoothness, and we use a fixed default $\tau=0.7$.
This objective yields a closed-form solution
\begin{equation}
p_t^*=\frac{1}{\mathcal{Z}}\exp\left(\frac{\kappa_t}{\tau}\right), \quad
\mathcal{Z}=\sum_{j=1}^{T}\exp\left(\frac{\kappa_j}{\tau}\right).
\end{equation}
We convert $\mathbf{p}^*$ into integer budgets $\{b_t\}_{t=1}^{T}$ with $\sum_t b_t=B$ by rounding $Bp_t^*$ and assigning leftover tokens to the largest remainders (then clip $b_t$ to $[0,P]$ when needed), so high-curvature transitions receive more tokens without overly concentrating the budget.

\subsection{Dual-Anchor Spatial Token Selection}
\label{subsec:spatial_sampling}

Given the per-frame budget $b_t\ll P$, we select spatial tokens that preserve both diversity and saliency within each frame.
We compute a \textbf{Dual-Anchor} score for each token $\mathbf{x}_{t,i}$:
\begin{equation}
s_{t,i} =
\underbrace{\big(1-\cos(\mathbf{x}_{t,i},\bar{\mathbf{x}}_t)\big)}_{\text{Contextual Diversity Anchor}}
+ \underbrace{\mathcal{N}\big(\|\mathbf{x}_{t,i}\|_2\big)}_{\text{Feature Activation Anchor}},
\label{eq:spatial_score}
\end{equation}
where $\mathcal{N}(\cdot)$ denotes min--max normalization.

\textbf{Contextual Diversity Anchor.}
The first term favors tokens that deviate from the global frame context $\bar{\mathbf{x}}_t$, which often correspond to subjects, interactions, or locally distinctive regions, and helps avoid keeping many redundant background patches.
\textbf{Feature Activation Anchor.}
The second term uses the $\ell_2$-norm as a lightweight saliency cue; empirically, higher-norm ViT features tend to indicate stronger activations and visually informative patterns.

We form $\mathcal{S}_t$ by selecting the Top-$b_t$ tokens with the largest $s_{t,i}$ and keep their original spatial coordinates, which preserves on-grid positional alignment under MRoPE.
Computing $\{s_{t,i}\}$ is linear in $P$, and Top-$b_t$ selection can be implemented efficiently with partial selection in practice.

\subsection{Theoretical Insight on Spatio-temporal Coverage} \label{subsec:theory} 

\noindent \textbf{Temporal Allocation via Curvature.} V-CAST represents frame semantics as $\{\hat{\mathbf{x}}_t\}_{t=1}^{T}\subset\mathbb{S}^{D'-1}$ and allocates budgets based on direction changes. Let $\Delta_t=\hat{\mathbf{x}}_{t+1}-\hat{\mathbf{x}}_t$ and $\kappa_t=1-\cos(\Delta_{t-1},\Delta_t)$. Since $\cos(\mathbf{a},\mathbf{b})=\frac{\mathbf{a}^{\top}\mathbf{b}}{\|\mathbf{a}\|_2\|\mathbf{b}\|_2+\epsilon}$, larger $\kappa_t$ indicates a larger turn between adjacent directions, \textit{i.e.}, a more significant semantic transition. We route the global budget by \begin{equation} \max_{\mathbf{p}\in\Delta^{T-1}} \sum_{t=1}^{T} p_t\kappa_t + \tau\mathcal{H}(\mathbf{p}) \quad\Rightarrow\quad p_t^*=\frac{1}{\mathcal{Z}}\exp\!\left(\frac{\kappa_t}{\tau}\right), \end{equation} so frames with larger $\kappa_t$ receive exponentially more budget while $\tau>0$ prevents collapse to a few frames, which improves temporal coverage under tight budgets.

\noindent \textbf{Spatial Selection via Diversity--Saliency.} For each token, V-CAST uses \begin{equation} s_{t,i}=\big(1-\cos(\mathbf{x}_{t,i},\bar{\mathbf{x}}_t)\big)+\mathcal{N}(\|\mathbf{x}_{t,i}\|_2). \end{equation} Let $\tilde{\mathbf{x}}_{t,i}=\mathbf{x}_{t,i}/\|\mathbf{x}_{t,i}\|_2$ and $\hat{\mathbf{x}}_t=\bar{\mathbf{x}}_t/\|\bar{\mathbf{x}}_t\|_2$. Then \begin{equation} 1-\cos(\mathbf{x}_{t,i},\bar{\mathbf{x}}_t) =1-\cos(\tilde{\mathbf{x}}_{t,i},\hat{\mathbf{x}}_t) =\tfrac{1}{2}\|\tilde{\mathbf{x}}_{t,i}-\hat{\mathbf{x}}_t\|_2^2, \end{equation} which favors directionally diverse tokens around $\hat{\mathbf{x}}_t$, while $\mathcal{N}(\|\mathbf{x}_{t,i}\|_2)$ favors salient activations. Selecting Top-$b_t$ by $s_{t,i}$ thus preserves diverse yet informative spatial evidence, and pruning keeps all retained tokens on-grid, maintaining positional alignment under MRoPE.
\section{Experiments}
\label{sec:Experiments}

\subsection{Experimental Setting}
\label{subsec:Setting}

\noindent \textbf{Benchmarks.} Our evaluation conducts on several multimodal video understanding benchmarks covering videos of diverse durations and scenarios, including MVBench~\cite{li2024mvbench}, LongVideoBench~\cite{wu2024longvideobench}, MLVU~\cite{zhou2024mlvu}, VideoMME~\cite{fu2024videomme}, and EgoSchema~\cite{mangalam2023egoschema}. We adopt the LMMs-Eval framework~\cite{zhang2024lmms} for evaluation. Detailed information is provided in Appendix~\ref{sec:appendix/benchmarks}.

\noindent \textbf{Implementations.} We validate token compression on various state-of-the-art VideoLLMs, including Qwen3-VL~\cite{Qwen3-VL}, LLaVA-Video~\cite{zhang2024llava-video}, and LLaVA-OneVision~\cite{li2024llava-ov}. We follow the default settings and set LLaVA-OV to take 32 input frames and LLaVA-Video to take 64 input frames. All experiments are conducted on NVIDIA H100-80GB HBM3 GPUs. Appendix~\ref{sec:appendix/models} provides more details.

\noindent \textbf{Baselines.} We compare our V-CAST against representative token compression methods for VideoLLMs, including  VisionZip~\cite{yang2025visionzip}, FastVID~\cite{shen2025fastvid}, HoliTom~\cite{shao2025holitom}, VidCom$^2$~\cite{liu2025vidcom2} and FlashVID~\cite{fan2026flashvid}. We re-implement all baseline token compression methods under the same experimental settings.

\subsection{Main Comparisons}
\label{subsec:Comparison}
\begin{table*}[!t]

\caption{\textbf{Performance comparison with other baselines on Qwen3-VL-8B-Instruct across benchmarks under multiple settings.} ``Average'' shows the mean performance across benchmarks.}
\vspace{-1mm}
\resizebox{\textwidth}{!}{
\begin{tabular}{lccccccccc}
\toprule
\footnotesize
\multirow{2}{*}{\textbf{Methods}} &
\multirow{2}{*}{\textbf{MVBench}} &
\multirow{2}{*}{\makecell{\textbf{LongVideo}\\\textbf{Bench}}} &
\multirow{2}{*}{\textbf{MLVU}} &
\multicolumn{4}{c}{\textbf{VideoMME}} &
\multicolumn{2}{c}{\textbf{Average}} \\
& & & &
\textbf{Overall} & \textbf{Short} & \textbf{Medium} & \textbf{Long} &
\textbf{Score} & \textbf{\%} \\
\midrule

\multicolumn{10}{c}{\textbf{\textit{Max Input Frames=64}}} \\
\midrule
\textcolor[rgb]{ .502,  .502,  .502}{Qwen3-VL-8B-Instruct} &
\textcolor[rgb]{ .502,  .502,  .502}{69.2} &
\textcolor[rgb]{ .502,  .502,  .502}{62.8} &
\textcolor[rgb]{ .502,  .502,  .502}{68.9} &
\textcolor[rgb]{ .502,  .502,  .502}{66.9} &
\textcolor[rgb]{ .502,  .502,  .502}{78.8} &
\textcolor[rgb]{ .502,  .502,  .502}{66.2} &
\textcolor[rgb]{ .502,  .502,  .502}{55.8} &
\textcolor[rgb]{ .502,  .502,  .502}{67.0} &
\textcolor[rgb]{ .502,  .502,  .502}{100.0} \\
\midrule

\multicolumn{10}{l}{\textit{Retention Ratio=25\%}} \\
VisionZip \conf{CVPR'25} &64.8 &58.9 &63.4 &63.1 &73.7 &60.3 &\underline{55.4} &62.6 &93.4 \\
VidCom$^2$ \conf{EMNLP'25} &\underline{67.5} &59.6 &64.0 &\underline{64.9} &\underline{75.4} &\textbf{63.4} &\textbf{55.9} &\underline{64.0} &\underline{95.5} \\
FastVID \conf{NeurIPS'25} &66.8 &\underline{60.4} &\underline{65.0} &62.3 &73.9 &60.7 &52.2 &63.6 &94.9 \\
HoliTom \conf{NeurIPS'25} &64.8 &58.9 &63.4 &62.7 &74.4 &60.6 &53.2 &62.5 &93.3 \\
FlashVID \conf{ICLR'26} & OOM & OOM & OOM & OOM & OOM & OOM & OOM & - & - \\
\rowcolor[rgb]{0.992, 0.961, 0.973}
\textbf{V-CAST \conf{Ours}} &\textbf{68.4} &\textbf{61.2} &\textbf{65.4} &\textbf{65.1} &\textbf{76.8} &\underline{63.2} &55.3 &\textbf{65.0} &\textbf{97.0} \\
\midrule

\multicolumn{10}{l}{\textit{Retention Ratio=15\%}} \\
VisionZip \conf{CVPR'25} &62.0 &57.7 &61.4 &60.4 &70.3 &58.4 &52.6 &60.4 &90.1 \\
VidCom$^2$ \conf{EMNLP'25} &64.3 &57.4 &60.3 &\underline{62.9} &\underline{72.6} &\underline{61.3} &\textbf{54.7} &61.2 &91.3 \\
FastVID \conf{NeurIPS'25} &\underline{64.9} &\textbf{59.7} &\underline{63.3} &60.5 &72.2 &57.2 &52.1 &\underline{62.1} &\underline{92.7} \\
HoliTom \conf{NeurIPS'25} &62.7 &58.3 &61.8 &60.9 &72.3 &58.2 &52.2 &60.9 &90.9 \\
FlashVID \conf{ICLR'26} & OOM & OOM & OOM & OOM & OOM & OOM & OOM & - & - \\
\rowcolor[rgb]{0.992, 0.961, 0.973}
\textbf{V-CAST \conf{Ours}} &\textbf{66.3} &\underline{59.6} &\textbf{64.4} &\textbf{63.9} &\textbf{76.3} &\textbf{61.4} &\underline{53.8} &\textbf{63.6} &\textbf{94.9} \\
\midrule

\multicolumn{10}{c}{\textbf{\textit{Max Input Frames=32}}} \\
\midrule
\textcolor[rgb]{ .502,  .502,  .502}{Qwen3-VL-8B-Instruct} &
\textcolor[rgb]{ .502,  .502,  .502}{68.6} &
\textcolor[rgb]{ .502,  .502,  .502}{60.3} &
\textcolor[rgb]{ .502,  .502,  .502}{63.5} &
\textcolor[rgb]{ .502,  .502,  .502}{64.5} &
\textcolor[rgb]{ .502,  .502,  .502}{76.0} &
\textcolor[rgb]{ .502,  .502,  .502}{60.4} &
\textcolor[rgb]{ .502,  .502,  .502}{57.0} &
\textcolor[rgb]{ .502,  .502,  .502}{64.2} &
\textcolor[rgb]{ .502,  .502,  .502}{100.0} \\
\midrule

\multicolumn{10}{l}{\textit{Retention Ratio=25\%}} \\
VisionZip \conf{CVPR'25} &62.2 &56.7 &60.8 &60.1 &69.6 &56.7 &54.2 &60.0 &93.5 \\
VidCom$^2$ \conf{EMNLP'25} &67.0 &58.0 &60.6 &\underline{62.4} &\underline{72.1} &\underline{59.1} &\textbf{56.1} &62.0 &96.6 \\
FastVID \conf{NeurIPS'25} &67.3 &\underline{58.7} &60.7 &60.5 &72.0 &56.8 &52.7 &61.8 &96.3 \\
HoliTom \conf{NeurIPS'25} &63.0 &56.8 &61.2 &59.7 &71.4 &54.6 &53.1 &60.2 &93.8 \\
FlashVID \conf{ICLR'26} &\underline{67.5} &\textbf{58.8} &\underline{61.7} &62.3 &\textbf{74.4} &58.7 &53.9 &\underline{62.6} &\underline{97.5} \\
\rowcolor[rgb]{0.992, 0.961, 0.973}
\textbf{V-CAST \conf{Ours}} &\textbf{67.9} &58.2 &\textbf{63.5} &\textbf{63.5} &\textbf{74.4} &\textbf{60.2} &\underline{55.8} &\textbf{63.3} &\textbf{98.6} \\
\midrule

\multicolumn{10}{l}{\textit{Retention Ratio=15\%}} \\
VisionZip \conf{CVPR'25} &60.1 &56.2 &\underline{60.0} &58.2 &66.9 &54.9 &52.9 &58.6 &91.3 \\
VidCom$^2$ \conf{EMNLP'25} &64.2 &56.0 &57.7 &59.6 &68.9 &56.7 &53.3 &59.4 &92.5 \\
FastVID \conf{NeurIPS'25} &66.1 &57.1 &59.0 &58.3 &69.6 &54.3 &51.1 &60.1 &93.6 \\
HoliTom \conf{NeurIPS'25} &60.0 &55.8 &59.6 &58.3 &68.9 &54.7 &51.7 &58.4 &91.0 \\
FlashVID \conf{ICLR'26} &\textbf{66.5} &\textbf{57.8} &\underline{60.0} &\underline{61.4} &\textbf{72.4} &\underline{57.6} &\underline{54.1} &\underline{61.4} &\underline{95.6} \\
\rowcolor[rgb]{0.992, 0.961, 0.973}
\textbf{V-CAST \conf{Ours}} &\underline{66.2} &\underline{57.7} &\textbf{60.1} &\textbf{62.0} &\underline{71.9} &\textbf{58.3} &\textbf{55.9} &\textbf{61.5} &\textbf{95.8} \\
\bottomrule
\end{tabular}
}
\vspace{-3.5mm}
\label{tab:qwen_results}
\end{table*}

\noindent \textbf{Comparisons on Qwen3-VL-8B-Instruct.} Table~\ref{tab:qwen_results} compares V-CAST with existing methods on Qwen3-VL-8B-Instruct and highlights three key observations. \textbf{(i) On MRoPE-based Qwen3-VL, pruning tends to outperform merging:}
Pruning methods, such as VidCom$^2$~\cite{liu2025vidcom2} and V-CAST, consistently outperform token merging methods, such as HoliTom~\cite{shao2025holitom} and FastVID~\cite{shen2025fastvid}. This trend is expected because Qwen3-VL adopts a three-dimensional MRoPE for video, and merging shifts token positions away from their original spatio-temporal coordinates, which weakens spatio-temporal alignment during downstream video understanding. \textbf{(ii) Stronger gains with longer inputs:} V-CAST performs better under the 64-frame setting than under the 32-frame setting at the same retention ratio, since it removes temporally redundant tokens while preserving salient motion and event cues, thereby retaining rich spatio-temporal information for long-context reasoning. \textbf{(iii) Extreme compression robustness:} V-CAST remains robust under aggressive compression. For Max Input Frames$=32$, it achieves \textbf{63.3} at $R=25\%$ (preserving \textbf{98.6\%} of the uncompressed performance) and maintains \textbf{62.3} at $R=15\%$ (corresponding to \textbf{97.0\%} retention), while ranking first in both settings. Notably, FlashVID runs out of memory under 64-frame inputs, likely because its attention-guided compression is executed inside the LLM and incurs higher activation and attention overhead, which also hinders efficient FlashAttention~\cite{daoFlashAttention-2} execution in long-video settings.

\begin{table*}[!t]    

\caption{\textbf{Performance comparison with other baselines with LLaVA-OV-7B across different benchmarks.} We use the default 32-frame input setting.}
\vspace{-1mm}
\resizebox{\textwidth}{!}{
    \begin{tabular}{lccccccccc}
    \toprule
    \multirow{2}{*}{\textbf{Methods}} &
    \multirow{2}{*}{\textbf{MVBench}} &
    \multirow{2}{*}{\makecell{\textbf{LongVideo}\\\textbf{Bench}}} &
    \multirow{2}{*}{\textbf{MLVU}} &
    \multicolumn{4}{c}{\textbf{VideoMME}} &
    \multicolumn{2}{c}{\textbf{Average}} \\
        &  &  &  &
        \textbf{Overall} & \textbf{Short} & \textbf{Medium} & \textbf{Long} &
        \textbf{Score} & \textbf{\%} \\
    \midrule
    \textcolor[rgb]{ .502,  .502,  .502}{LLaVA-OV-7B} & \textcolor[rgb]{ .502,  .502,  .502}{58.3} & \textcolor[rgb]{ .502,  .502,  .502}{56.6} & \textcolor[rgb]{ .502,  .502,  .502}{63.1} & \textcolor[rgb]{ .502,  .502,  .502}{58.4} & \textcolor[rgb]{ .502,  .502,  .502}{69.9} & \textcolor[rgb]{ .502,  .502,  .502}{56.7} & \textcolor[rgb]{ .502,  .502,  .502}{48.8} & \textcolor[rgb]{ .502,  .502,  .502}{59.1} & \textcolor[rgb]{ .502,  .502,  .502}{100.0} \\
    \midrule

    \multicolumn{10}{l}{\textit{Retention Ratio=25\%}} \\
    FastV \conf{ECCV'24} & 55.5 & 53.3 & 59.6 & 55.3 & 65.0 & 53.8 & 47.0 & 55.9 & 94.6 \\
    PDrop \conf{CVPR'25} & 55.3 & 51.3 & 57.1 & 55.5 & 64.7 & 53.1 & 48.7 & 54.8 & 92.7 \\
    SparseVLM \conf{ICML'25} & 56.4 & 53.9 & 60.7 & 57.3 & 68.4 & 55.2  & 48.1 & 57.1 & 96.6 \\
    VisionZip \conf{CVPR'25} & 56.9 & 56.0 & \underline{62.9} & 58.0 & 68.9 & \underline{57.4} & 47.6 & \underline{58.5} & \underline{99.0} \\
    PruneVid \conf{ACL'25} & 55.7 & 55.1 & \textbf{63.4} & 57.0 & 68.8 & 54.4 & 47.7 & 57.8 & 97.8 \\
    FrameFusion \conf{ICCV'25} & 56.0 & 54.8 & 61.7 & 57.5 & 68.2 & 55.7 & 48.6 & 57.5 & 97.3 \\
    FastVID \conf{NeurIPS'25} & 56.5 & \underline{56.3} & 60.9 & 58.3 & \underline{69.4} & \textbf{58.2} & 47.2 & 58.0 & 98.1 \\
    VidCom$^2$ \conf{EMNLP'25} & \underline{57.0} & 55.4 & 62.8 & \underline{58.4} & 69.3 & 56.3 & \textbf{49.4} & 58.4 & 98.8 \\
    \rowcolor[rgb]{0.992, 0.961, 0.973}
    \textbf{V-CAST \conf{Ours}} & \textbf{57.4} & \textbf{56.4} & \underline{62.9} & \textbf{58.6} & \textbf{70.7} & 56.0 & \underline{49.1} & \textbf{58.8} & \textbf{99.5} \\
    \bottomrule
    \end{tabular}%
    }
    \vspace{-2.6mm}
  \label{tab:llava_ov_results}%
\end{table*}
\begin{table*}[!t]    

\caption{\textbf{Performance comparison with other baselines with LLaVA-Video-7B across different benchmarks.} We use the default 64-frame setting.} 
\vspace{-1mm} 
\resizebox{\textwidth}{!}{ 
    \begin{tabular}{lccccccccc} 
    \toprule 
    \multirow{2}{*}{\textbf{Methods}} &  
    \multirow{2}{*}{\textbf{MVBench}} &  
    \multirow{2}{*}{\makecell{\textbf{LongVideo}\\\textbf{Bench}}} & 
    \multirow{2}{*}{\textbf{MLVU}} &  
    \multicolumn{4}{c}{\textbf{VideoMME}} &  
    \multicolumn{2}{c}{\textbf{Average}} \\ 
        &   &   &   &  
        \textbf{Overall} & \textbf{Short} & \textbf{Medium} & \textbf{Long} &  
        \textbf{Score} & \textbf{\%} \\ 
    \midrule 
    \textcolor[rgb]{ .502,  .502,  .502}{LLaVA-Video-7B} & 
    \textcolor[rgb]{ .502,  .502,  .502}{60.4} & 
    \textcolor[rgb]{ .502,  .502,  .502}{58.9} & 
    \textcolor[rgb]{ .502,  .502,  .502}{67.3} & 
    \textcolor[rgb]{ .502,  .502,  .502}{64.4} & 
    \textcolor[rgb]{ .502,  .502,  .502}{77.3} & 
    \textcolor[rgb]{ .502,  .502,  .502}{62.4} & 
    \textcolor[rgb]{ .502,  .502,  .502}{53.4} & 
    \textcolor[rgb]{ .502,  .502,  .502}{62.8} & 
    \textcolor[rgb]{ .502,  .502,  .502}{100.0} \\ 
    \midrule 

    \multicolumn{10}{l}{\textit{Retention Ratio=25\%}} \\ 
    FastV \conf{ECCV'24} &52.1  &54.8  &57.8  &58.6  &68.7  &58.4  &48.7  &55.8 &88.9 \\ 
    SparseVLM \conf{ICML'25} &55.4 &54.2 &58.9 &60.1 &71.1 &59.1 &50.1 &57.2 &91.1 \\ 
    VisionZip \conf{CVPR'25} &57.9  &56.3  &\textbf{62.6}  &62.5  &73.6  &\textbf{62.3}  &51.9  &\underline{59.8} &\underline{95.2} \\ 
    HoliTom \conf{NeurIPS'25} &\textbf{58.4}  &\underline{57.1}  &60.5  &\textbf{63.0}  &\textbf{74.6}  &\textbf{62.3}  &\underline{52.1}  &\underline{59.8} &\underline{95.2} \\ 
    VidCom$^2$ \conf{EMNLP'25} &57.0  &\underline{57.1}  &58.7  &61.7  &73.0  &\underline{61.7}  &50.0  &58.6 &93.4 \\ 
    \rowcolor[rgb]{0.992, 0.961, 0.973} 
    \textbf{V-CAST \conf{Ours}} &\underline{58.0} &\textbf{57.3} &\underline{61.6}  &\underline{62.7}  &\underline{74.0} &61.2 &\textbf{52.4}  &\textbf{59.9} &\textbf{95.4} \\ 
    \bottomrule 
    \end{tabular}%
    } 
    \vspace{-4.5mm} 
  \label{tab:llava_video_results}%
\end{table*}

\noindent \textbf{Comparisons on LLaVA-OV-7B and LLaVA-Video-7B.} Table~\ref{tab:llava_ov_results} and Table~\ref{tab:llava_video_results} compare V-CAST with baseline methods on the LLaVA series VideoLLMs and show that it delivers consistent improvements without any training. Notably, these two backbones rely on standard RoPE rather than the three-dimensional MRoPE in Qwen3-VL, which confirms that V-CAST is broadly compatible with common positional encoding designs. Under $R=25\%$, V-CAST achieves an average score of \textbf{58.8} on LLaVA-OV-7B and \textbf{59.9} on LLaVA-Video-7B, preserving \textbf{99.5\%} and \textbf{95.4\%} of the original performance, respectively. These results indicate that the same budgeting and token selection rule transfers reliably across VideoLLM backbones and generalizes well to videos of different lengths.

\noindent \textbf{Comparisons on Larger VideoLLM Qwen3-VL-32B-Instruct.} Table~\ref{tab:qwen3vl_32b} compares V-CAST with existing methods on Qwen3-VL-32B-Instruct and confirms its scalability on larger models. Under $R=25\%$, V-CAST preserves \textbf{98.2\%} of the original performance and achieves the best average score among all compressed baselines, outperforming the second-best VidCom$^2$~\cite{liu2025vidcom2} by \textbf{1.3\%} on average. Unlike FlashVID~\cite{fan2026flashvid}, which relies on attention scores for token compression, V-CAST compresses tokens based on the visual content itself, remaining compatible with FlashAttention~\cite{daoFlashAttention-2} and avoiding the extra activation and attention overhead of attention-guided compression. As a result, V-CAST provides plug-and-play acceleration on large models, showing that it transfers reliably to larger-capacity VideoLLMs and remains practical for long-context inference.

\noindent \textbf{Comparisons on MoE-based VideoLLM Qwen3-VL-30B-A3B-Instruct.}
Table~\ref{tab:qwen3vl_moe} compares V-CAST with baseline methods on Qwen3-VL-30B-A3B-Instruct and demonstrates its effectiveness under large-scale settings. Notably, the vanilla model and FlashVID~\cite{fan2026flashvid} encounter OOM errors under 64-frame inputs, while V-CAST enables stable and efficient inference. Under $R=25\%$, V-CAST achieves the best average score of \textbf{68.1} across all benchmarks. 
V-CAST shows clear advantages on long-video benchmarks, achieving the best results on LongVideoBench~\cite{wu2024longvideobench} with \textbf{66.6} and on VideoMME (Long)~\cite{fu2024videomme} with \textbf{59.8}. These results indicate that V-CAST scales well to MoE-based VideoLLMs and better preserves long-range spatio-temporal information under compression.

\begin{table*}[!t]    

\caption{\textbf{Performance comparison with other baselines on the larger model Qwen3-VL-32B-Instruct across different benchmarks.} We use 32-frame input setting.}
\vspace{-1mm}
\resizebox{\textwidth}{!}{
    \begin{tabular}{lccccccccc}
    \toprule
    \footnotesize
    \multirow{2}{*}{\textbf{Methods}} & 
    \multirow{2}{*}{\textbf{MVBench}} & 
    \multirow{2}{*}{\makecell{\textbf{LongVideo}\\\textbf{Bench}}} &
    \multirow{2}{*}{\textbf{MLVU}} & 
    \multicolumn{4}{c}{\textbf{VideoMME}} & 
    \multicolumn{2}{c}{\textbf{Average}} \\
        &  &  &  & 
        \textbf{Overall} & \textbf{Short} & \textbf{Medium} & \textbf{Long} & 
        \textbf{Score} & \textbf{\%} \\
    \midrule

    \textcolor[rgb]{ .502,  .502,  .502}{Qwen3-VL-32B-Instruct} & \textcolor[rgb]{ .502,  .502,  .502}{73.2} & \textcolor[rgb]{ .502,  .502,  .502}{62.4} & \textcolor[rgb]{ .502,  .502,  .502}{66.1} & \textcolor[rgb]{ .502,  .502,  .502}{69.3} & \textcolor[rgb]{ .502,  .502,  .502}{78.3} & \textcolor[rgb]{ .502,  .502,  .502}{67.1} & \textcolor[rgb]{ .502,  .502,  .502}{62.4} & \textcolor[rgb]{ .502,  .502,  .502}{67.8} & \textcolor[rgb]{ .502,  .502,  .502}{100.0} \\
    \midrule

    \multicolumn{10}{l}{\textit{Retention Ratio=25\%}} \\
    VisionZip \conf{CVPR'25} &62.9   &60.0   &62.2   &64.8   &72.8   &63.1   &58.6   &62.5 &92.2 \\
    VidCom$^2$ \conf{EMNLP'25} &70.2   &60.4   &\textbf{64.9}   &\underline{67.0}   &\underline{75.6}   &\underline{64.3}   &\textbf{61.2}   &\underline{65.6} &\underline{96.8} \\
    FastVID \conf{NeurIPS'25} &\underline{71.0}   &\underline{60.9}   &64.5   &65.1   &75.1   &62.6   &57.7   &65.4 &96.5 \\
    HoliTom \conf{NeurIP'S25} &62.3   &60.2   &63.2   &64.9   &74.2   &61.2   &59.1 &62.6 &92.3 \\
    FlashVID \conf{ICLR'26} & OOM   & OOM   & OOM   & OOM   & OOM   & OOM   & OOM & -  & -  \\
    \rowcolor[rgb]{0.992, 0.961, 0.973}   
    \textbf{V-CAST \conf{Ours}} &\textbf{71.8}   &\textbf{61.5}   &\underline{64.7}  &\textbf{68.1}   &\textbf{77.7}   &\textbf{65.9}   &\underline{60.8}   &\textbf{66.5} &\textbf{98.1} \\
    \bottomrule
    \end{tabular}
    }
    \vspace{-2.9mm}
  \label{tab:qwen3vl_32b}
\end{table*}

\begin{table*}[!t]    

\caption{\textbf{Performance comparison with other baselines on the MoE-based model Qwen3-VL-30B-A3B-Instruct across different benchmarks.} We use 64-frame input setting.}
\vspace{-1mm}
\resizebox{\textwidth}{!}{
    \begin{tabular}{lcccccccc}
    \toprule
    \footnotesize
    \multirow{2}{*}{\textbf{Methods}} & 
    \multirow{2}{*}{\textbf{MVBench}} & 
    \multirow{2}{*}{\makecell{\textbf{LongVideo}\\\textbf{Bench}}} &
    \multirow{2}{*}{\textbf{MLVU}} & 
    \multicolumn{4}{c}{\textbf{VideoMME}} & 
    \multirow{2}{*}{\textbf{Average Score}} \\
        &  &  &  & 
        \textbf{Overall} & \textbf{Short} & \textbf{Medium} & \textbf{Long} &  \\
    
    \midrule
    \textcolor[rgb]{ .502,  .502,  .502}{Qwen3-VL-30B-A3B-Instruct} & \textcolor[rgb]{ .502,  .502,  .502}{OOM} & \textcolor[rgb]{ .502,  .502,  .502}{OOM} & \textcolor[rgb]{ .502,  .502,  .502}{OOM} & \textcolor[rgb]{ .502,  .502,  .502}{OOM} & \textcolor[rgb]{ .502,  .502,  .502}{OOM} & \textcolor[rgb]{ .502,  .502,  .502}{OOM} & \textcolor[rgb]{ .502,  .502,  .502}{OOM} & \textcolor[rgb]{ .502,  .502,  .502}{-} \\
    \midrule

    \multicolumn{9}{l}{\textit{Retention Ratio=25\%}} \\
    VisionZip \conf{CVPR'25} & 61.3 & 62.4 & 64.1 & 62.7 & 70.7 & 61.3 & 56.0 & 62.6 \\
    VidCom$^2$ \conf{EMNLP'25} & \textbf{69.7} & \underline{65.5} & \textbf{68.5} & \underline{68.1} & \textbf{79.1} & \textbf{68.0} & \underline{58.9} & \underline{68.0} \\
    FastVID \conf{NeurIPS'25} & 68.2 & 60.7 & 64.8 & 62.4 & 70.3 & 61.6 & 55.2 & 64.0 \\
    HoliTom \conf{NeurIPS'25} & 66.0 & 63.9 & 64.7 & 64.6 & 73.6 & \underline{63.9} & 56.2 & 64.8 \\
    FlashVID \conf{ICLR'26} & OOM & OOM & OOM & OOM & OOM & OOM & OOM & - \\
    \rowcolor[rgb]{0.992, 0.961, 0.973}
    \textbf{V-CAST \conf{Ours}} & \underline{69.2} & \textbf{66.6} & \underline{68.2} & \textbf{68.2} & \underline{78.9} & 65.9 & \textbf{59.8} & \textbf{68.1} \\
    \bottomrule
    \end{tabular}
}
\vspace{-2.2mm}
\label{tab:qwen3vl_moe}
\end{table*}

\begin{figure*}[!t]
    \centering
    \includegraphics[width=\linewidth]{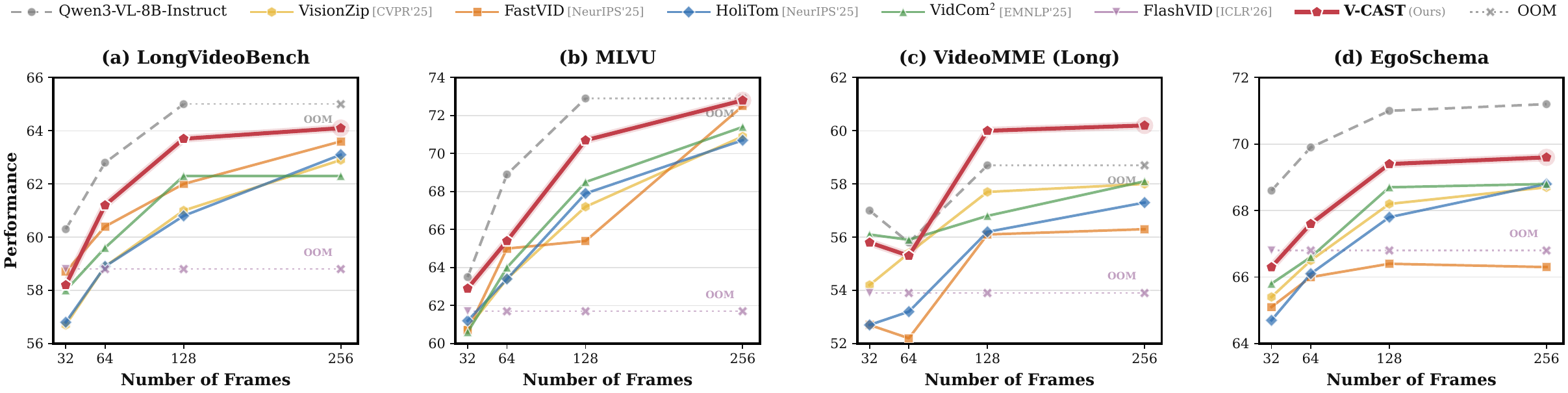}
    \vspace{-5mm}
    \caption{\textbf{Consistent gains with more frames.} Performance trends on LongVideoBench, MLVU, VideoMME (Long), and EgoSchema as input frames increase. V-CAST improves accuracy and scales to longer inputs, while some baselines show limited gains or OOM failures at larger frame counts.}
    \label{fig:frame_scaling}
     \vspace{-4.9mm}
\end{figure*}

\noindent \textbf{Comparisons on the Impact of Numbers of Frames.} Figure~\ref{fig:frame_scaling} evaluates how performance changes as we increase the maximum number of input frames on four typical long-video benchmarks, including LongVideoBench~\cite{wu2024longvideobench}, MLVU~\cite{zhou2024mlvu}, EgoSchema~\cite{mangalam2023egoschema}, and VideoMME (Long)~\cite{fu2024videomme}. In general, using more frames improves accuracy for both the vanilla model and most compression methods,  indicating that longer temporal contexts can provide more helpful evidence. However, FlashVID~\cite{fan2026flashvid} runs out of memory beyond 64 frames because it relies on explicit attention scores for token compression, while the vanilla model and other training-free compressors do not exhibit this issue at the same scale, suggesting that practical compression should remain compatible with efficient attention kernels. When the maximum input reaches 256 frames, V-CAST shows clear advantages over prior methods, outperforming the second-best VidCom$^2$~\cite{liu2025vidcom2} by \textbf{+1.5} points on average across benchmarks and widening the margin over segment-based methods~\cite{shao2025holitom,shen2025fastvid} and uniform compression~\cite{yang2025visionzip}. These results indicate that V-CAST better preserves continuous spatio-temporal information coverage as the context grows, leading to more robust long-video understanding overall.

\begin{figure*}[!t]
    \centering
    \includegraphics[width=\linewidth]{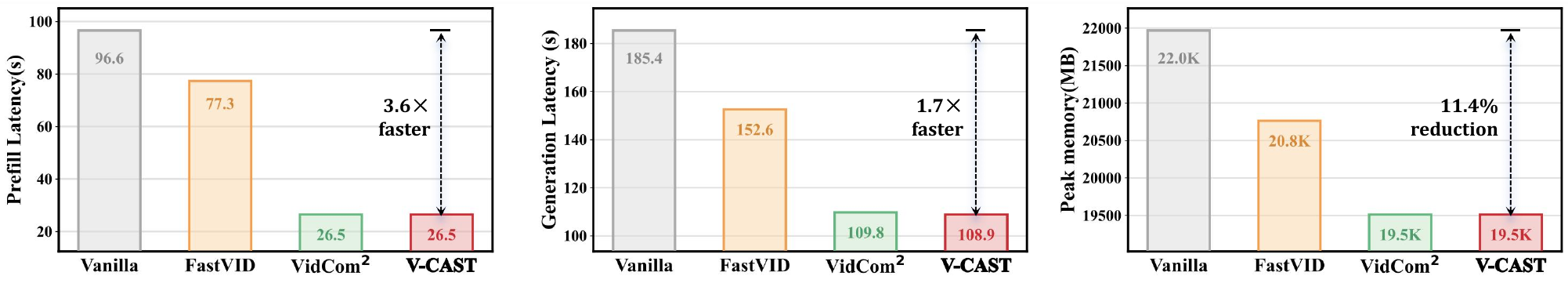}
    \vspace{-5.5mm}
    \caption{\textbf{Efficiency comparison on LLaVA-OneVision-7B.} We compare Vanilla, FastVID, VidCom$^2$, and V-CAST on Prefill Latency, Total Latency, and Peak Memory. }
    \label{fig:llava_ov_efficiency}
    \vspace{-2mm}
\end{figure*}

\noindent \textbf{Comparisons on Inference Efficiency.} Table~\ref{tab:efficiency_comparison} compares inference efficiency of token compression methods on Qwen3-VL-8B-Instruct. V-CAST achieves the best overall efficiency, reducing prefilling latency by \textbf{51.2\%} and peak memory by \textbf{13.0\%} while maintaining strong throughput. In contrast, FlashVID~\cite{fan2026flashvid} incurs much higher memory overhead, increasing peak memory by \textbf{84.0\%} compared to the uncompressed model, which makes it prone to out-of-memory failures for long-context inference. Figure~\ref{fig:llava_ov_efficiency} shows similarly strong gains on LLaVA-OneVision-7B: V-CAST achieves \textbf{3.6$\times$} faster prefilling, \textbf{1.7$\times$} faster generation, and \textbf{11.4\%} peak-memory reduction. These results jointly demonstrate that V-CAST provides the most practical inference efficiency for VideoLLMs.

Extensive comparisons show that V-CAST performs well across diverse benchmarks and video tasks. It generalizes to VideoLLMs of different architectures and scales, and remains robust across a wide range of context lengths. Meanwhile, V-CAST delivers a strong accuracy--efficiency trade-off, achieving state-of-the-art performance with the best inference efficiency.

\begin{table*}[!t]    

\definecolor{greenrightcolor}{RGB}{0,144,81}
\definecolor{redrightcolor}{RGB}{255,0,0}

\newcommand{\downtinya}[1]{{\!\textcolor{greenrightcolor}{\tiny{#1}}}}
\newcommand{\uptiny}[1]{{\!\textcolor{redrightcolor}{\tiny{#1}}}}

\caption{\textbf{Efficiency comparisons.} Qwen3-VL-8B-Instruct on VideoMME at $R=25\%$ (max input frames = 32). ``Prefill Latency'': prompt-to-first-token; ``LLM Generation Latency'': decoding time (first-to-last-token); ``Total Latency'': end-to-end wall-clock time in our experimental setup; ``Throughput'': samples per second.}
\vspace{-1mm}
\resizebox{\textwidth}{!}{
    \begin{tabular}{lcccccc}
    \toprule
    \textbf{Methods} &
    \textbf{Prefilling} $\downarrow$ &
    \textbf{LLM Generation} $\downarrow$ &
    \textbf{Total Latency} $\downarrow$ &
    \textbf{GPU Peak} $\downarrow$ &
    \textbf{Throughput} $\uparrow$ &
    \textbf{Performance} $\uparrow$ \\
    & \textbf{Latency (s)} & \textbf{Latency (s)} & \textbf{(s)} & \textbf{Memory (MB)} & \textbf{(item/s)} & \\
    \midrule

    \textcolor[rgb]{ .502,  .502,  .502}{Qwen3-VL-8B-Instruct} &
    \textcolor[rgb]{ .502,  .502,  .502}{243.6} &
    \textcolor[rgb]{ .502,  .502,  .502}{280.9} &
    \textcolor[rgb]{ .502,  .502,  .502}{1440.4} &
    \textcolor[rgb]{ .502,  .502,  .502}{22478.0} &
    \textcolor[rgb]{ .502,  .502,  .502}{1.87} &
    \textcolor[rgb]{ .502,  .502,  .502}{64.5} \\
    \midrule

    Random & 116.9 \downtinya{(↓52.0\%)} & 147.3 \downtinya{(↓47.6\%)} & 1271.9 \downtinya{(↓11.7\%)} & 19547.5 \downtinya{(↓13.0\%)} & 2.12 \downtinya{(1.13$\times$)} & 61.5 \downtinya{(↓3.0)} \\
    \midrule

    VisionZip \conf{CVPR'25} & 155.6 \downtinya{(↓36.1\%)} & 190.5 \downtinya{(↓32.2\%)} & 1276.8 \downtinya{(↓11.4\%)} & \underline{19974.7} \downtinya{(↓11.2\%)} & 2.11 \downtinya{(1.13$\times$)} & 60.1 \downtinya{(↓4.4)} \\
    VidCom$^2$ \conf{EMNLP'25} & \underline{119.0} \downtinya{(↓51.1\%)} & 150.2 \downtinya{(↓46.5\%)} & 1271.0 \downtinya{(↓11.8\%)} & \textbf{19547.5} \downtinya{(↓13.0\%)} & 2.12 \downtinya{(1.13$\times$)} & \underline{62.4} \downtinya{(↓2.1)} \\
    FastVID \conf{NeurIPS'25} & 120.6 \downtinya{(↓50.5\%)} & 153.6 \downtinya{(↓45.3\%)} & 1317.6 \downtinya{(↓8.5\%)} & \textbf{19547.5} \downtinya{(↓13.0\%)} & 2.05 \downtinya{(1.10$\times$)} & 60.5 \downtinya{(↓4.0)} \\
    HoliTom \conf{NeurIPS'25} & 134.3 \downtinya{(↓44.9\%)} & 160.8 \downtinya{(↓42.8\%)} & \underline{1261.8} \downtinya{(↓12.4\%)} & \underline{19974.7} \downtinya{(↓11.2\%)} & \underline{2.14} \downtinya{(1.14$\times$)} & 59.7 \downtinya{(↓4.8)} \\
    FlashVID \conf{ICLR'26} & 146.6 \downtinya{(↓39.8\%)} & 175.0 \downtinya{(↓37.7\%)} & 1308.7 \downtinya{(↓9.1\%)} & 41363.6 \uptiny{(↑84.0\%)} & 2.06 \downtinya{(1.10$\times$)} & 62.3 \downtinya{(↓2.2)} \\
    \rowcolor[rgb]{0.992, 0.961, 0.973}
    \textbf{V-CAST \conf{Ours}} &
    \textbf{118.8} \downtinya{(↓51.2\%)} & \textbf{149.9} \downtinya{(↓46.7\%)} &
    \textbf{1245.1} \downtinya{(↓13.6\%)} & \textbf{19547.5} \downtinya{(↓13.0\%)} & \textbf{2.17} \downtinya{(1.16$\times$)} & \textbf{63.5} \downtinya{(↓1.0)} \\

    \bottomrule
    \end{tabular}
}
\vspace{-2mm}
\label{tab:efficiency_comparison}
\end{table*}

\subsection{Ablation Study and Analysis}
\label{subsec:ablation}

\begin{figure*}[!t]
  \centering
  \includegraphics[width=\linewidth]{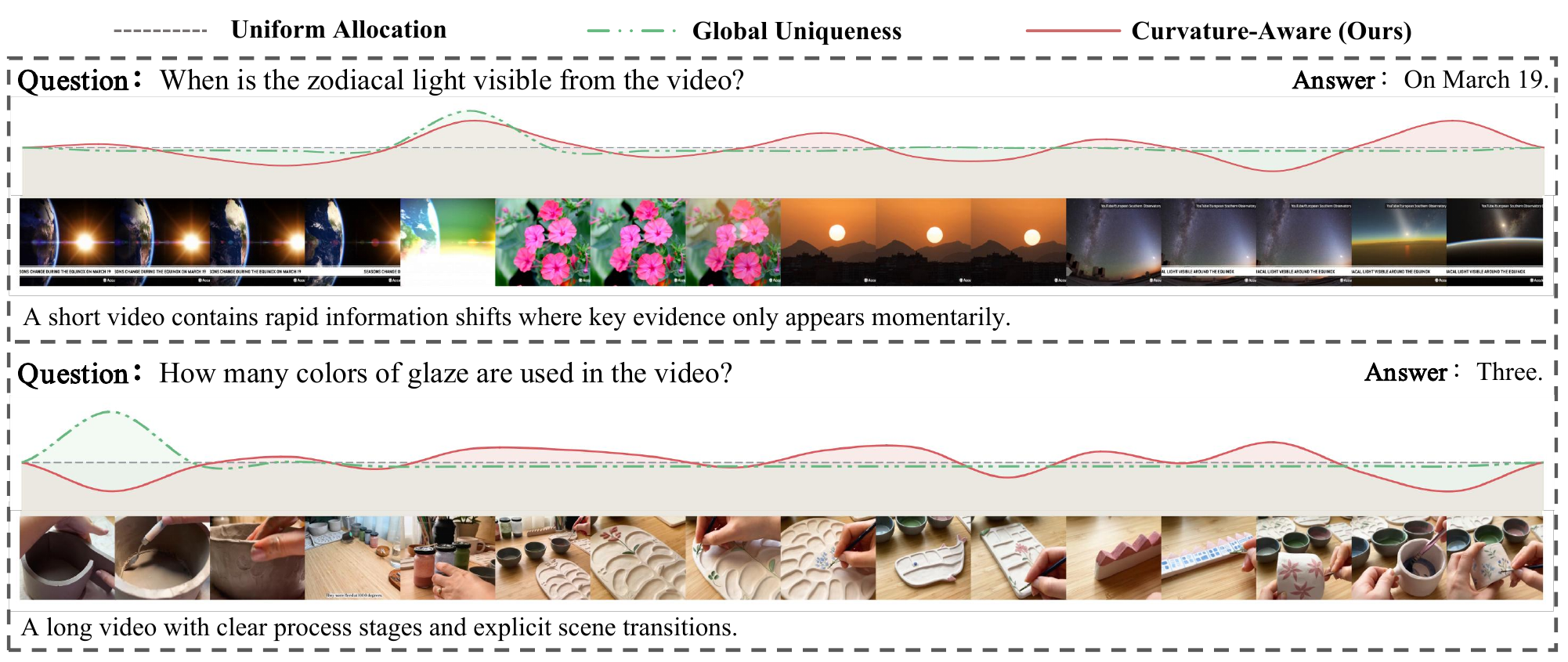}
  \vspace{-4mm}
  \caption{\textbf{Visualization of frame budget allocation.} We compare \textcolor[rgb]{0.557,0.576,0.596}{Uniform allocation}, \textcolor[rgb]{0.400,0.722,0.502}{Global Uniqueness} (VidCom$^2$~\cite{liu2025vidcom2}), and \textcolor[rgb]{0.851,0.353,0.373}{Curvature-Aware allocation} (V-CAST) under a $R=25\%$. A higher curve indicates a larger per-frame token budget.}
  \label{fig:temporal_allocation}
  \vspace{-3.5mm}
\end{figure*}

\noindent \textbf{Ablation on Temporal Allocation.} Table~\ref{tab:ablation_temporal_32_64} evaluates temporal routing strategies under $R=25\%$ with Max Input Frames$=32$ and 64 respectively, while Figure~\ref{fig:temporal_allocation} visualizes their per-frame budgets on representative videos. Uniform Allocation assigns the same budget to every frame, implicitly assuming homogeneous information density; the nearly flat curves show that it cannot emphasize brief yet decisive moments, leading to weaker results. Global Uniqueness (VidCom$^2$~\cite{liu2025vidcom2}-style) improves over Uniform by allocating more tokens to globally distinctive frames, which explains its modest gains and more adaptive curves. However, because it relies on coarse frame-level uniqueness, it can still over-concentrate budgets on a few frames and under-allocate transitional segments where evidence appears only momentarily. In contrast, Curvature-Aware allocation (V-CAST) routes budgets to high-curvature transitions, producing sharper yet more targeted peaks in Figure~\ref{fig:temporal_allocation} and achieving the best overall scores under both settings.

\begin{table*}[!t]
    \centering
    \caption{\textbf{Ablation on Temporal Allocation.} Evaluating temporal routing metrics on Qwen3-VL-8B at $R=25\%$ under different maximum input frames.}
    \vspace{-1.5mm}

    \begin{subtable}[t]{0.49\textwidth}
        \centering
        \caption{Max Input Frames$=32$.}
        \vspace{-1.5mm}
        \resizebox{\linewidth}{!}{
        \begin{tabular}{lccccc}
            \toprule
            \footnotesize
            \multirow{2}{*}{\textbf{Temporal Allocation}} &
            \multirow{2}{*}{\textbf{MLVU}} &
            \multicolumn{3}{c}{\textbf{VideoMME}} &
            \textbf{Average} \\
            \cmidrule(lr){3-5}
            & & \textbf{Overall} & \textbf{Short} & \textbf{Medium} & \textbf{Score} \\
            \midrule
            \textcolor[rgb]{.502,.502,.502}{Qwen3-VL-8B-Instruct} &
            \textcolor[rgb]{.502,.502,.502}{63.5} &
            \textcolor[rgb]{.502,.502,.502}{64.5} &
            \textcolor[rgb]{.502,.502,.502}{76.0} &
            \textcolor[rgb]{.502,.502,.502}{60.4} &
            \textcolor[rgb]{.502,.502,.502}{64.0} \\
            \midrule
            Uniform Allocation & 61.6 & 62.9 & 73.9 & \underline{60.0} & 62.3 \\
            Random Allocation  & 62.0 & 62.6 & 73.7 & 59.9 & 62.3 \\
            Global Uniqueness  & \underline{62.1} & \underline{63.0} & \underline{74.1} & 59.9 & \underline{62.6} \\
            \rowcolor[rgb]{0.992,0.961,0.973}
            \textbf{Curvature-Aware \conf{Ours}}  & \textbf{62.7} & \textbf{63.5} & \textbf{74.4} & \textbf{60.2} & \textbf{63.1} \\
            \bottomrule
        \end{tabular}}
        \label{tab:ablation_temporal_32}
    \end{subtable}\hfill
    \begin{subtable}[t]{0.49\textwidth}
        \centering
        \caption{Max Input Frames$=64$.}
        \vspace{-1.5mm}
        \resizebox{\linewidth}{!}{
        \begin{tabular}{lccccc}
            \toprule
            \footnotesize
            \multirow{2}{*}{\textbf{Temporal Allocation}} &
            \multirow{2}{*}{\textbf{MLVU}} &
            \multicolumn{3}{c}{\textbf{VideoMME}} &
            \textbf{Average} \\
            \cmidrule(lr){3-5}
            & & \textbf{Overall} & \textbf{Short} & \textbf{Medium} & \textbf{Score} \\
            \midrule
            \textcolor[rgb]{.502,.502,.502}{Qwen3-VL-8B-Instruct} &
            \textcolor[rgb]{.502,.502,.502}{68.9} &
            \textcolor[rgb]{.502,.502,.502}{66.9} &
            \textcolor[rgb]{.502,.502,.502}{78.8} &
            \textcolor[rgb]{.502,.502,.502}{66.2} &
            \textcolor[rgb]{.502,.502,.502}{67.9} \\
            \midrule
            Uniform Allocation & \underline{65.2} & 64.7 & \underline{76.6} & 62.7 & 65.0 \\
            Random Allocation  & 64.7 & 64.4 & 75.9 & \underline{62.9} & 64.6 \\
            Global Uniqueness  & 65.1 & \underline{64.8} & 76.2 & \underline{62.9} & \underline{65.0} \\
            \rowcolor[rgb]{0.992,0.961,0.973}
            \textbf{Curvature-Aware \conf{Ours}} & \textbf{65.4} & \textbf{65.1} & \textbf{76.8} & \textbf{63.2} & \textbf{65.3} \\
            \bottomrule
        \end{tabular}}
        \label{tab:ablation_temporal_64}
    \end{subtable}

    \vspace{-2.5mm}
    \label{tab:ablation_temporal_32_64}
\end{table*}

\begin{wraptable}{r}{0.575\textwidth}
    \centering
    \vspace{-4.6mm}
    \caption{\textbf{Ablation on Spatial Selection.} Evaluating spatial selection anchors on Qwen3-VL-8B at $R=25\%$.}
    \vspace{-0.5mm}
    \resizebox{\linewidth}{!}{
        \begin{tabular}{lccccccc}
        \toprule
        \footnotesize
        \multirow{2}{*}{\textbf{Spatial Selection}} &
        \textbf{MVBench} &
        \textbf{MLVU} &
        \multicolumn{3}{c}{\textbf{VideoMME}} &
        \textbf{Average} \\
        \cmidrule(lr){4-6}
        &  &  &
        \textbf{Overall} & \textbf{Short} & \textbf{Medium} &
        \textbf{Score} \\
        \midrule
        \textcolor[rgb]{ .502, .502, .502}{Qwen3-VL-8B-Instruct} &
        \textcolor[rgb]{ .502, .502, .502}{68.6} &
        \textcolor[rgb]{ .502, .502, .502}{63.5} &
        \textcolor[rgb]{ .502, .502, .502}{64.5} &
        \textcolor[rgb]{ .502, .502, .502}{76.0} &
        \textcolor[rgb]{ .502, .502, .502}{60.4} &
        \textcolor[rgb]{ .502, .502, .502}{65.5} \\
        \midrule
        Contextual Diversity Only & \underline{67.5} & \underline{61.6} & 61.2 & \underline{71.3} & 57.1 & 63.4 \\
        Feature Activation Only   & 63.0 & 58.6 & \underline{61.4} & 70.0 & \underline{57.2} & 61.0 \\
        \rowcolor[rgb]{0.992, 0.961, 0.973}
        \textbf{Dual-Anchor \conf{Ours}}      & \textbf{67.9} & \textbf{62.7} & \textbf{63.5} & \textbf{74.4} & \textbf{60.2} & \textbf{64.7} \\
        \bottomrule
        \end{tabular}
    }
    \vspace{-1em}
    \label{tab:ablation_spatial}
\end{wraptable}

\noindent \textbf{Ablation on Spatial Selection.} Table~\ref{tab:ablation_spatial} compares spatial token selection variants under the same settings. Contextual Diversity Only encourages diversity but may suppress important, context-consistent evidence, while Feature Activation Only captures salient responses but can keep redundant high-activation patches and miss complementary cues. Dual-Anchor combines both signals and achieves the best overall performance by preserving diverse and informative evidence within each frame.

\begin{figure*}[!t]
  \centering
   \includegraphics[width=\linewidth]{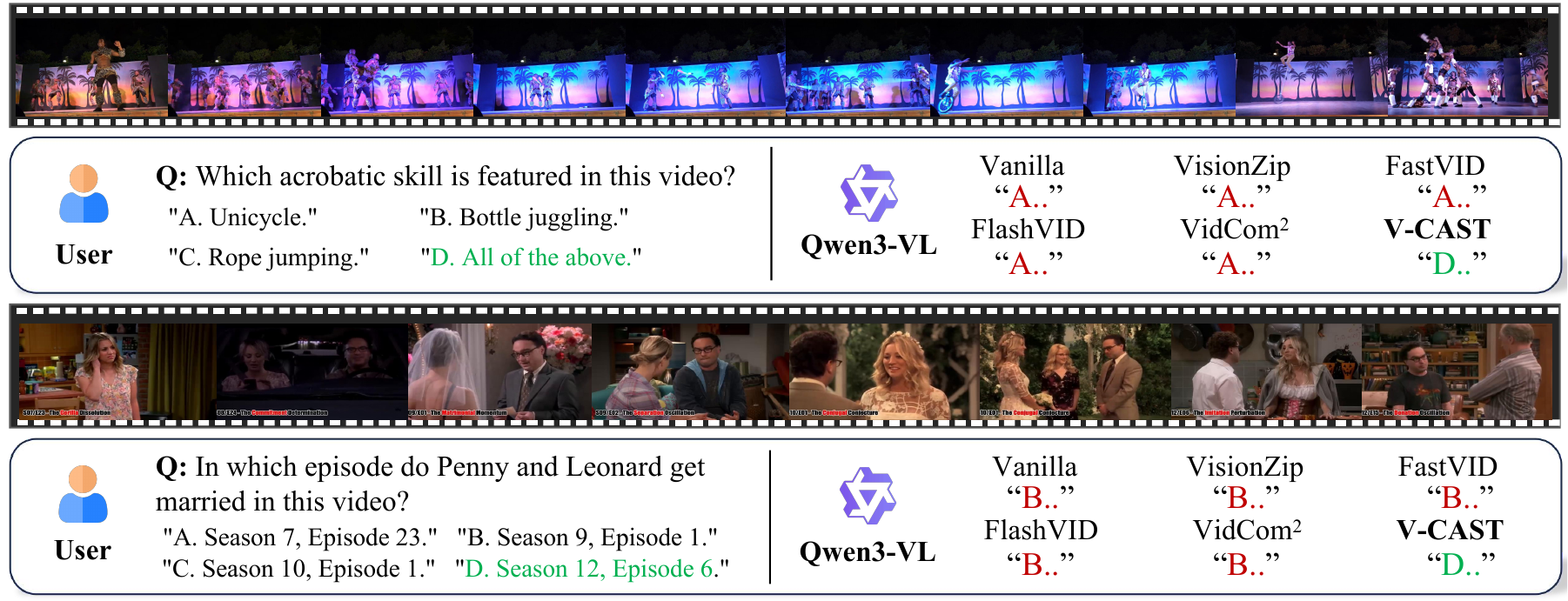}
   \vspace{-4.5mm}
    \caption{\textbf{Qualitative comparison.} V-CAST highlights task-critical moments and yields correct answers where baselines and even the vanilla model fail.}
   \label{fig:badcase}
   \vspace{-4mm}
\end{figure*}

\begin{wrapfigure}[12]{r}{0.5\textwidth}
    \centering
    \vspace{-4.3mm}
    \includegraphics[width=\linewidth]{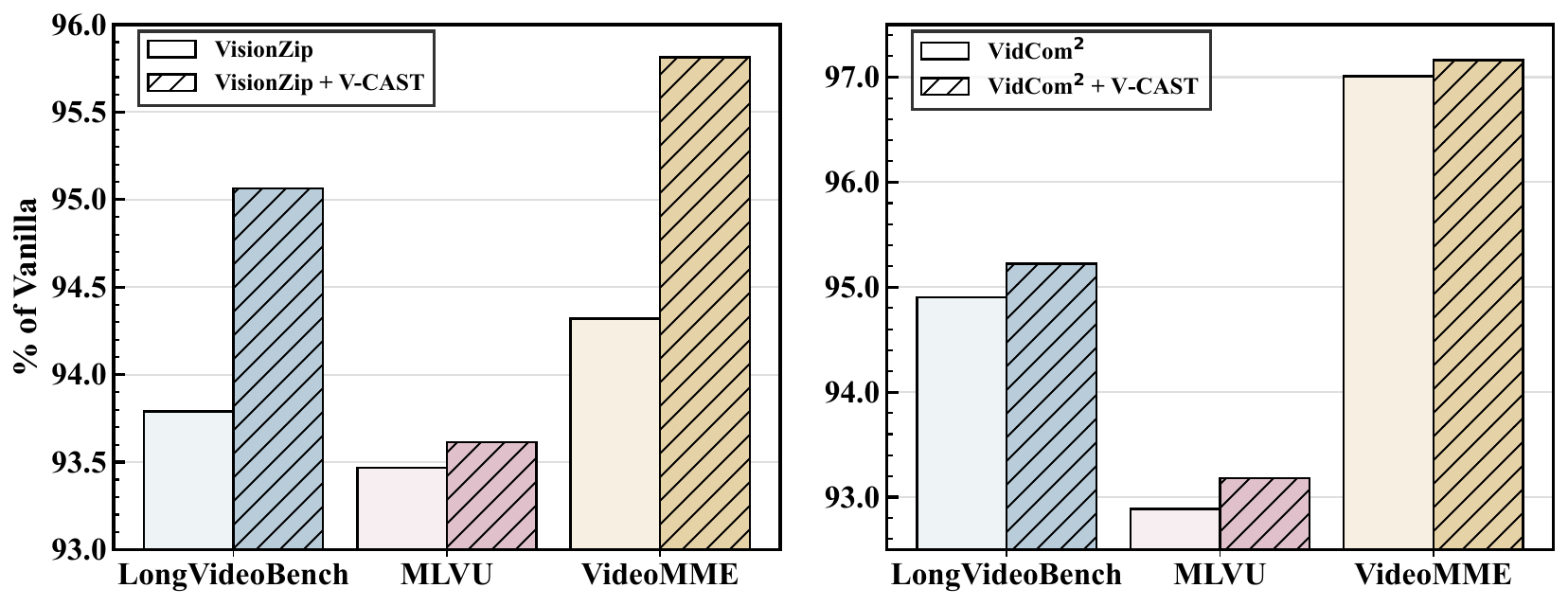}
    \caption{\textbf{Broad applicability of V-CAST budgeting on other frame compression methods.}
    ``+V-CAST'' denotes equipping the original method with our curvature-aware budgeting strategy.}
    \label{fig:ablation_budget_others}
    \vspace{-2mm}
\end{wrapfigure}

\noindent \textbf{Broad Applicability of Curvature-Aware Budgeting.}
Figure~\ref{fig:ablation_budget_others} shows that our curvature-aware budgeting can be readily and effectively integrated into other compression methods. Under the same $R=25\%$, adding curvature-aware budgeting consistently improves the performance of VisionZip~\cite{yang2025visionzip} and VidCom$^2$~\cite{liu2025vidcom2} across different tasks. The gains are more evident for VisionZip, suggesting that adaptive temporal budget allocation is especially beneficial when the original method lacks explicit temporal importance modeling. For VidCom$^2$, the improvements are smaller but consistent, as expected since it already accounts for frame uniqueness during compression. These results suggest that our budgeting is broadly compatible with different compression paradigms and can serve as a generic temporal allocation strategy.

\noindent \textbf{Qualitative Analysis.} Figure~\ref{fig:badcase} compares different methods under $R=25\%$. By adaptively allocating more tokens to transition windows and retaining salient complementary patches, V-CAST keeps decisive evidence with nearby context, enabling correct answers. In contrast, uniform allocation or coarse frame-wise budgeting either spreads tokens over redundant regions or skips brief evidence frames, which can obscure the key cue and lead to incorrect reasoning.

\section{Conclusion}
\label{sec:Conclusion}

In this work, we revisit token compression for VideoLLMs under tight budgets and identify two critical issues: discontinuous spatio-temporal information coverage caused by coarse allocation or segmentation, and spatio-temporal misalignment introduced by token merging under MRoPE. Based on these insights, we propose V-CAST, a training-free and plug-and-play token pruning framework that casts compression as an optimal trajectory approximation problem. V-CAST combines curvature-guided temporal allocation with dual-anchor spatial selection to preserve continuous and well-aligned spatio-temporal evidence while keeping retained tokens at their original coordinates. Extensive experiments across various multiple VideoLLMs and benchmarks confirm that V-CAST achieves state-of-the-art performance and efficiency for video understanding.

\bibliographystyle{plain}
\bibliography{main}


\appendix


\newpage
\clearpage
\appendix

In this appendix, we provide additional details on the benchmarks used in our evaluation in Section~\ref{sec:appendix/benchmarks}, the baseline VideoLLMs in Section~\ref{sec:appendix/models}, supplementary experimental results in Section~\ref{sec:appendix/results}, formal algorithmic details of V-CAST in Section~\ref{sec:appendix/algorithm_vcast}, and further qualitative comparisons in Section~\ref{sec:appendix/more_qualitative}.

\section{Benchmark Details}
\label{sec:appendix/benchmarks}

This section provides a brief overview of the video understanding benchmarks used in our evaluation:

\begin{itemize}

    \item \textbf{MVBench}~\cite{li2024mvbench} assesses 20 distinct temporal tasks. It employs a static-to-dynamic method to generate questions from image datasets. The benchmark covers diverse perspectives from first-person to third-person. It requires models to perform complex reasoning beyond single-frame perception.
    
    \item \textbf{LongVideoBench}~\cite{wu2024longvideobench} targets long-context comprehension with 3,763 videos. It features a novel referring reasoning task for precise temporal localization. The dataset includes original or transcribed subtitles. This allows for the evaluation of interleaved multimodal inputs across 17 categories.
    
    \item \textbf{MLVU}~\cite{zhou2024mlvu} includes videos lasting 3 minutes to 2 hours. It evaluates 9 tasks such as action counting and temporal ordering. The benchmark requires models to process global summaries and fine-grained details. It highlights the challenges of maintaining long-term dependencies.
    
    \item \textbf{VideoMME}~\cite{fu2024videomme} consists of 900 high-quality videos across 6 primary domains. It integrates audio tracks and subtitles to test comprehensive multi-modal perception. The questions cover 30 fine-grained subfields. Content is categorized into short, medium, and long subsets for multi-scale analysis.
    
    \item \textbf{EgoSchema}~\cite{mangalam2023egoschema} is a diagnostic benchmark for long-form egocentric video understanding. It contains over 5,000 human-curated multiple-choice questions derived from the Ego4D dataset. Each task involves a three-minute video clip with long temporal context. It evaluates reasoning over complex human activities from a first-person perspective.
\end{itemize}

\section{Model Details}
\label{sec:appendix/models}

This section summarizes the VideoLLMs evaluated in our experiments:

\begin{itemize}

    \item \textbf{LLaVA-OneVision}~\cite{li2024llava-ov} establishes a unified framework for single-image, multi-image, and video tasks. It treats videos as interleaved sequences of visual tokens. This design facilitates seamless knowledge transfer from image-based pre-training to video understanding. The model demonstrates exceptional zero-shot performance across diverse temporal tasks.
    
    \item \textbf{LLaVA-Video}~\cite{zhang2024llava-video} initializes from the LLaVA-OneVision image-stage checkpoints. It undergoes fine-tuning on the LLaVA-Video-178K dataset. This large-scale synthetic corpus covers detailed captioning and complex reasoning. By integrating the SigLIP vision encoder with the Qwen2 language model, it achieves outstanding performance across benchmarks.

    \item \textbf{Qwen3-VL}~\cite{Qwen3-VL} represents the latest advancement in the Qwen-VL series. It enhances the visual perception system to support higher-resolution inputs and more efficient token compression. The model features an upgraded LLM backbone with improved reasoning logic and multilingual capabilities. It excels at fine-grained visual grounding and complex instruction following in long-duration video streams.

\end{itemize}

\section{More Experimental Results}
\label{sec:appendix/results}

\noindent \textbf{Performance under Higher Retention Ratio.}
Table~\ref{tab:qwen_0.35_results} evaluates V-CAST under a budget of $R=35\%$. V-CAST achieves an average score of 65.8 (98.2\%) with 64 frames and 63.4 (98.8\%) with 32 frames, showing that curvature-aware selection effectively identifies critical tokens as the budget increases. While FlashVID encounters OOM errors at 64 frames, V-CAST enables robust inference across different sequence lengths due to its attention-agnostic pruning. V-CAST consistently matches or surpasses the strongest baseline, suggesting that routing tokens toward semantic transitions can be more effective than global frame uniqueness for preserving spatio-temporal continuity under constrained budgets.

\begin{table*}[!t]

\caption{\textbf{Performance comparison with other baselines on Qwen3-VL-8B-Instruct under a retention ratio of $35\%$.} ``Average'' denotes the mean performance across benchmarks.}

\vspace{-2mm}
\resizebox{\textwidth}{!}{
\begin{tabular}{lccccccccc}
\toprule
\footnotesize
\multirow{2}{*}{\textbf{Methods}} &
\multirow{2}{*}{\textbf{MVBench}} &
\multirow{2}{*}{\makecell{\textbf{LongVideoBench}}} &
\multirow{2}{*}{\textbf{MLVU}} &
\multicolumn{4}{c}{\textbf{VideoMME}} &
\multicolumn{2}{c}{\textbf{Average}} \\
& & & &
\textbf{Overall} & \textbf{Short} & \textbf{Medium} & \textbf{Long} &
\textbf{Score} & \textbf{\%} \\
\midrule

\multicolumn{10}{c}{\textbf{\textit{Max Input Frames=64}}} \\
\midrule
\textcolor[rgb]{ .502,  .502,  .502}{Qwen3-VL-8B-Instruct} &
\textcolor[rgb]{ .502,  .502,  .502}{69.2} &
\textcolor[rgb]{ .502,  .502,  .502}{62.8} &
\textcolor[rgb]{ .502,  .502,  .502}{68.9} &
\textcolor[rgb]{ .502,  .502,  .502}{66.9} &
\textcolor[rgb]{ .502,  .502,  .502}{78.8} &
\textcolor[rgb]{ .502,  .502,  .502}{66.2} &
\textcolor[rgb]{ .502,  .502,  .502}{55.8} &
\textcolor[rgb]{ .502,  .502,  .502}{67.0} &
\textcolor[rgb]{ .502,  .502,  .502}{100.0} \\
\midrule

\multicolumn{10}{l}{\textit{Retention Ratio=35\%}} \\
VisionZip \conf{CVPR'25} & 65.7 & 58.9 & 64.3 & 64.3 & 74.9 & 62.2 & \textbf{55.9} & 63.3 & 94.5 \\
VidCom$^2$ \conf{EMNLP'25} & \textbf{68.9} & \underline{61.0} & 65.5 & \textbf{66.0} & \underline{77.3} & \underline{64.9} & \underline{55.8} & \underline{65.4} & \underline{97.6} \\
FastVID \conf{NeurIPS'25} & 66.5 & 60.7 & \underline{66.1} & 63.0 & 74.2 & 60.3 & 54.4 & 64.1 & 95.7 \\
HoliTom \conf{NeurIPS'25} & 65.8 & 60.4 & 64.4 & 64.0 & 75.8 & 61.8 & 54.3 & 63.6 & 95.0 \\
FlashVID \conf{ICLR'26} & OOM & OOM & OOM & OOM & OOM & OOM & OOM & - & - \\
\rowcolor[rgb]{0.992, 0.961, 0.973}
\textbf{V-CAST \conf{Ours}} & \underline{68.5} & \textbf{61.6} & \textbf{67.1} & \textbf{66.0} & \textbf{77.6} & \textbf{65.6} & 55.0 & \textbf{65.8} & \textbf{98.2} \\
\midrule

\multicolumn{10}{c}{\textbf{\textit{Max Input Frames=32}}} \\
\midrule
\textcolor[rgb]{ .502,  .502,  .502}{Qwen3-VL-8B-Instruct} &
\textcolor[rgb]{ .502,  .502,  .502}{68.6} &
\textcolor[rgb]{ .502,  .502,  .502}{60.3} &
\textcolor[rgb]{ .502,  .502,  .502}{63.5} &
\textcolor[rgb]{ .502,  .502,  .502}{64.5} &
\textcolor[rgb]{ .502,  .502,  .502}{76.0} &
\textcolor[rgb]{ .502,  .502,  .502}{60.4} &
\textcolor[rgb]{ .502,  .502,  .502}{57.0} &
\textcolor[rgb]{ .502,  .502,  .502}{64.2} &
\textcolor[rgb]{ .502,  .502,  .502}{100.0} \\
\midrule

\multicolumn{10}{l}{\textit{Retention Ratio=35\%}} \\
VisionZip \conf{CVPR'25} & 64.4 & 56.8 & 61.3 & 60.8 & 70.9 & 57.0 & 54.6 & 60.8 & 94.8 \\
VidCom$^2$ \conf{EMNLP'25} & \underline{68.1} & 58.6 & 62.0 & \underline{62.9} & 73.6 & \underline{59.9} & \underline{55.1} & 62.9 & 98.0 \\
FastVID \conf{NeurIPS'25} & 67.5 & \underline{58.7} & 62.3 & 61.9 & 73.1 & 59.1 & 53.6 & 62.6 & 97.5 \\
HoliTom \conf{NeurIPS'25} & 65.0 & 57.8 & 62.0 & 61.3 & 71.9 & 57.9 & 54.0 & 61.5 & 95.8 \\
FlashVID \conf{ICLR'26} & 67.9 & \textbf{59.2} & \underline{62.4} & 62.7 & \textbf{75.2} & 59.6 & 53.3 & \underline{63.1} & \underline{98.3} \\
\rowcolor[rgb]{0.992, 0.961, 0.973}
\textbf{V-CAST \conf{Ours}} & \textbf{68.3} & 58.5 & \textbf{63.1} & \textbf{63.9} & \underline{74.7} & \textbf{60.8} & \textbf{56.3} & \textbf{63.5} & \textbf{98.8} \\
\bottomrule
\end{tabular}
}
\vspace{-2mm}
\label{tab:qwen_0.35_results}
\end{table*}

\noindent \textbf{Hyper-parameter Sensitivity Analysis}
We further analyze the sensitivity of V-CAST to two key hyper-parameters in the main experiments: the temporal budget temperature $\tau$ (default: 0.7) and the spatial anchor weights (default: 1.0, 1.0).
Table~\ref{tab:ablation_full_suppl} (a) shows that V-CAST remains stable across $\tau$ choices, with the average score varying only from 65.0 to 65.3 as $\tau$ changes from 0.6 to 0.9. Among the tested values, $\tau=0.7$ gives the best trade-off, achieving the highest VideoMME overall, short, medium, and long scores, while matching the best MLVU result up to rounding. In contrast, a larger temperature ($\tau=0.9$) makes the allocation more uniform and slightly weakens focus on decisive frames. Overall, a moderate temperature best balances emphasis on critical moments and temporal context.
Table~\ref{tab:ablation_full_suppl} (b) shows that V-CAST remains robust under different anchor weight settings. The balanced choice $(1.0, 1.0)$ achieves the best average score of 65.3, suggesting that the two anchors provide complementary cues for spatial selection. Overall, V-CAST is not highly sensitive to the exact weight choice, while a balanced configuration yields the most consistent performance.

\begin{table*}[!h]
    \centering
    \caption{\textbf{Sensitivity analysis on V-CAST hyper-parameters.} We evaluate (a) temporal temperature $\tau$ and (b) spatial selection weights on Qwen3-VL-8B-Instruct at $R=25\%$.}
    \begin{subtable}[t]{0.48\textwidth}
        \centering
        \caption{Ablation on Temperature $\tau$.}
        \resizebox{\linewidth}{!}{
        \begin{tabular}{lcccccc}
            \toprule
            \footnotesize
            \multirow{2}{*}{\textbf{$\tau$}} &
            \multirow{2}{*}{\textbf{MLVU}} &
            \multicolumn{4}{c}{\textbf{VideoMME}} &
            \multirow{2}{*}{\textbf{Avg Score}} \\
            \cmidrule(lr){3-6}
            & & \textbf{Overall} & \textbf{Short} & \textbf{Medium} & \textbf{Long} & \\
            \midrule
            \textcolor[rgb]{.502,.502,.502}{Vanilla} &
            \textcolor[rgb]{.502,.502,.502}{68.9} &
            \textcolor[rgb]{.502,.502,.502}{66.9} &
            \textcolor[rgb]{.502,.502,.502}{78.8} &
            \textcolor[rgb]{.502,.502,.502}{66.2} &
            \textcolor[rgb]{.502,.502,.502}{55.8} &
            \textcolor[rgb]{.502,.502,.502}{67.9} \\
            \midrule
            0.6 & 65.3 & \underline{64.8} & 76.2 & \underline{63.1} & \underline{55.1} & \underline{65.1} \\
            \rowcolor[rgb]{0.992, 0.961, 0.973}
            0.7 & \textbf{65.4} & \textbf{65.1} & \textbf{76.8} & \textbf{63.2} & \textbf{55.3} & \textbf{65.3} \\
            0.9 & \textbf{65.4} & 64.6 & \underline{76.1} & 62.8 & 55.0 & 65.0 \\
            \bottomrule
        \end{tabular}}
    \end{subtable}\hfill
    \begin{subtable}[t]{0.48\textwidth}
        \centering
        \caption{Ablation on Anchor Weights.}
        \vspace{-1mm}
        \resizebox{\linewidth}{!}{
        \begin{tabular}{lcccccc}
            \toprule
            \footnotesize
            \multirow{2}{*}{\textbf{Weight}} &
            \multirow{2}{*}{\textbf{MLVU}} &
            \multicolumn{4}{c}{\textbf{VideoMME}} &
            \multirow{2}{*}{\textbf{Avg Score}} \\
            \cmidrule(lr){3-6}
            & & \textbf{Overall} & \textbf{Short} & \textbf{Medium} & \textbf{Long} & \\
            \midrule
            \textcolor[rgb]{.502,.502,.502}{Vanilla} &
            \textcolor[rgb]{.502,.502,.502}{68.9} &
            \textcolor[rgb]{.502,.502,.502}{66.9} &
            \textcolor[rgb]{.502,.502,.502}{78.8} &
            \textcolor[rgb]{.502,.502,.502}{66.2} &
            \textcolor[rgb]{.502,.502,.502}{55.8} &
            \textcolor[rgb]{.502,.502,.502}{67.9} \\
            \midrule
            (1.3, 0.7) & 65.0 & \underline{64.9} & 76.0 & 63.0 & \textbf{55.7} & 65.0 \\
            (1.1, 0.9) & 65.0 & 64.7 & 75.6 & \textbf{63.2} & 55.2 & 64.9 \\
            \rowcolor[rgb]{0.992, 0.961, 0.973}
            (1.0, 1.0) & \underline{65.4} & \textbf{65.1} & \underline{76.8} & \textbf{63.2} & \underline{55.3} & \textbf{65.3} \\
            (0.9, 1.1) & \textbf{65.7} & 64.7 & \textbf{77.0} & 62.6 & 54.7 & \underline{65.2} \\
            \bottomrule
        \end{tabular}}
    \end{subtable}
    \label{tab:ablation_full_suppl}
\end{table*}

\section{Algorithm Details of V-CAST}
\label{sec:appendix/algorithm_vcast}
Algorithm~\ref{alg:vcast} summarizes V-CAST. It first allocates the global token budget across frames according to semantic curvature, and then selects the top-$b_t$ spatial tokens in each frame using the dual-anchor diversity-saliency score.

\section{More Qualitative Comparisons}

\label{sec:appendix/more_qualitative}

\begin{figure*}[!t]
  \centering
   \includegraphics[width=\linewidth]{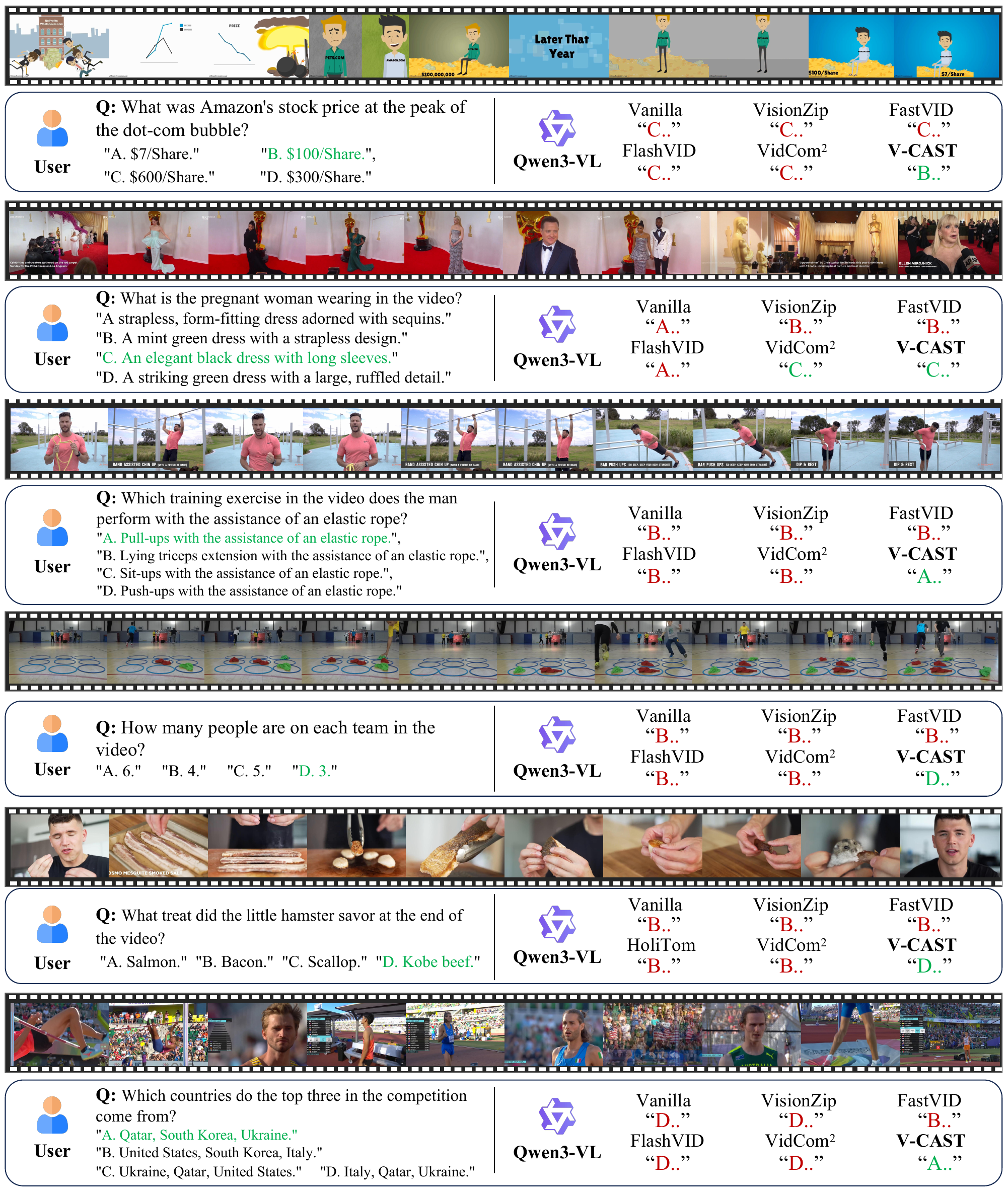}
   \vspace{-6mm}
    \caption{\textbf{Additional qualitative comparisons.} V-CAST preserves task-critical moments and correctly answers challenging cases where baseline compression methods, and occasionally even the vanilla model, miss decisive visual evidence.}
   \label{fig:more_badcase}
   \vspace{-4mm}
\end{figure*}

Figure~\ref{fig:more_badcase} presents additional qualitative comparisons among V-CAST, various token compression baselines, and the vanilla VideoLLM. These cases cover a diverse range of challenging video understanding scenarios, including rapid action shifts, subtle semantic transitions, and complex long-term reasoning. The examples show that V-CAST more effectively identifies and preserves brief but decisive moments by dynamically routing token budgets to critical event boundaries. Unlike existing methods that often dilute fleeting evidence under tight budgets, V-CAST highlights and preserves these task-critical moments. Consequently, V-CAST yields correct answers in complex cases where conventional compression baselines, and sometimes even the vanilla model, fail to capture the decisive visual cues.

\begin{algorithm}[h]
\caption{V-CAST: Curvature-Guided Spatio-Temporal Token Pruning}
\label{alg:vcast}
\begin{algorithmic}[1]
\Require Visual tokens $\mathbf{X}=\{\mathbf{x}_{t,i}\}\in\mathbb{R}^{T\times P\times D'}$, retention ratio $r$, temperature $\tau$, minimum per-frame budget $k_{\min}$
\Ensure Pruned visual token sequence $\tilde{\mathbf{X}}^{\flat}$

\State $B \gets \lfloor rTP \rfloor$
\For{$t=1,\dots,T$}
    \State $\bar{\mathbf{x}}_t \gets \frac{1}{P}\sum_{i=1}^{P}\mathbf{x}_{t,i}$, \quad
    $\hat{\mathbf{x}}_t \gets \bar{\mathbf{x}}_t / (\|\bar{\mathbf{x}}_t\|_2+\epsilon)$
\EndFor

\For{$t=1,\dots,T$}
    \State $\Delta_t \gets \hat{\mathbf{x}}_{t+1}-\hat{\mathbf{x}}_t$
    \State $\kappa_t \gets 1-\cos(\Delta_{t-1}, \Delta_t)$ \Comment{boundary indices are clamped}
\EndFor

\State $p_t \gets \exp(\kappa_t/\tau)\big/\sum_{j=1}^{T}\exp(\kappa_j/\tau)$
\State Convert $\{p_t\}_{t=1}^{T}$ into integer budgets $\{b_t\}_{t=1}^{T}$ such that $\sum_t b_t=B$ and $b_t\in[k_{\min},P]$

\For{$t=1,\dots,T$}
    \For{$i=1,\dots,P$}
        \State $s_{t,i} \gets \big(1-\cos(\mathbf{x}_{t,i},\bar{\mathbf{x}}_t)\big) + \mathcal{N}(\|\mathbf{x}_{t,i}\|_2)$
    \EndFor
    \State $\mathcal{S}_t \gets \operatorname{TopB}\big(\{\mathbf{x}_{t,i}\}_{i=1}^{P}, \{s_{t,i}\}_{i=1}^{P}, b_t\big)$
\EndFor

\State \Return $\tilde{\mathbf{X}}^{\flat} \gets \operatorname{Concat}(\mathcal{S}_1,\dots,\mathcal{S}_T)$
\end{algorithmic}
\end{algorithm}

\end{document}